\documentclass[letterpaper, 10 pt, conference]{ieeeconf}  

\IEEEoverridecommandlockouts                              
\usepackage{graphicx}
\usepackage{array}
\usepackage{amsmath}
\usepackage{tikz}
\usepackage{amssymb}
\usepackage{mathtools}  
\usepackage{booktabs}   
\usepackage{multirow}   
\usepackage{cite}
\usepackage{subcaption}  

\usepackage{xurl}                     
\usepackage[hidelinks]{hyperref}      
\usepackage{textcomp}                 

\title{\LARGE \bf
FastViDAR: Real-Time Omnidirectional Depth Estimation\\
via Alternative Hierarchical Attention
}

\author{Hangtian ZHAO$^{1}$, Xiang Chen$^{2}$, Yizhe Li$^{3}$, Qianhao Wang$^{4}$, Haibo Lu, and Fei Gao$^{4*}$
\thanks{*Corresponding author: Fei Gao}
\thanks{$^{1}$Author is with the University of Science and Technology of China.
        {\tt\small htzhao@mail.ustc.edu.cn}}%
\thanks{$^{2}$Author is with East China Normal University.
        {\tt\small 71285901012@stu.ecnu.edu.cn}}%
\thanks{$^{3}$Author is with Xidian University.
        {\tt\small liyizhe666@stu.xidian.edu.cn}}%
\thanks{$^{4}$Authors are with the FAST Lab, Zhejiang University.
        {\tt\small \{wangqianhao, feigao\}@zju.edu.cn}}%
}

\begin{document}

\maketitle
\thispagestyle{empty}
\pagestyle{empty}

\begin{abstract}
        In this paper we propose FastViDAR, a novel framework that takes four fisheye camera inputs and produces a full $360^\circ$ depth map along with per-camera depth, fusion depth, and confidence estimates. Our main contributions are: (1) We introduce Alternative Hierarchical Attention (AHA) mechanism that efficiently fuses features across views through separate intra-frame and inter-frame windowed self-attention, achieving cross-view feature mixing with reduced overhead. (2) We propose a novel ERP fusion approach that projects multi-view depth estimates to a shared equirectangular coordinate system to obtain the final fusion depth. (3) We generate ERP image-depth pairs using HM3D and 2D3D-S datasets for comprehensive evaluation, demonstrating competitive zero-shot performance on real datasets while achieving up to 20 FPS on NVIDIA Orin NX embedded hardware. Project page: \href{https://3f7dfc.github.io/FastVidar/}{https://3f7dfc.github.io/FastVidar/}
        \end{abstract}

\section{Introduction}

Fast and reliable omnidirectional depth is crucial for robotics and autonomous driving. Active sensors (e.g., LiDAR) provide accurate $360^\circ$ depth but are costly and power-hungry, whereas multi-camera rigs with fisheye lenses offer a practical alternative. A four-camera rig with ultra-wide field of view (FOV) ($>180^\circ$) covers the full sphere, but inferring a consistent, accurate, and efficient depth map from these views remains challenging. Classic extensions of stereo to fisheye rely on spherical epipolar geometry and plane sweeping with volumetric cost aggregation, which often hampers real-time deployment and assumes perfect inter-camera extrinsics~\cite{CasOmniMVS2024, Won_2019}. Recent monocular approaches generalize well across cameras by factoring out intrinsics; notably Depth Any Camera~\cite{DAC2025} converts inputs (perspective/fisheye/panorama) to a common equirectangular projection (ERP) to achieve zero-shot metric depth~\cite{DAC2025}. However, single-image methods cannot leverage multi-view geometry nor estimate inter-camera parameters. In parallel, transformer-based multi-view models such as VGGT~\cite{Wang2025VGGT} aggregate cross-view information with alternating local/global attention and predict camera parameters and dense depth in a feed-forward manner, suggesting self-attention~\cite{Vaswani2017Attention} can replace heavy cost volumes—though achieving real-time inference on autonomous mobile robot platforms remains challenging.

\begin{figure}[t]
  \includegraphics[width=0.48\textwidth,trim={110 120 400 42},clip]{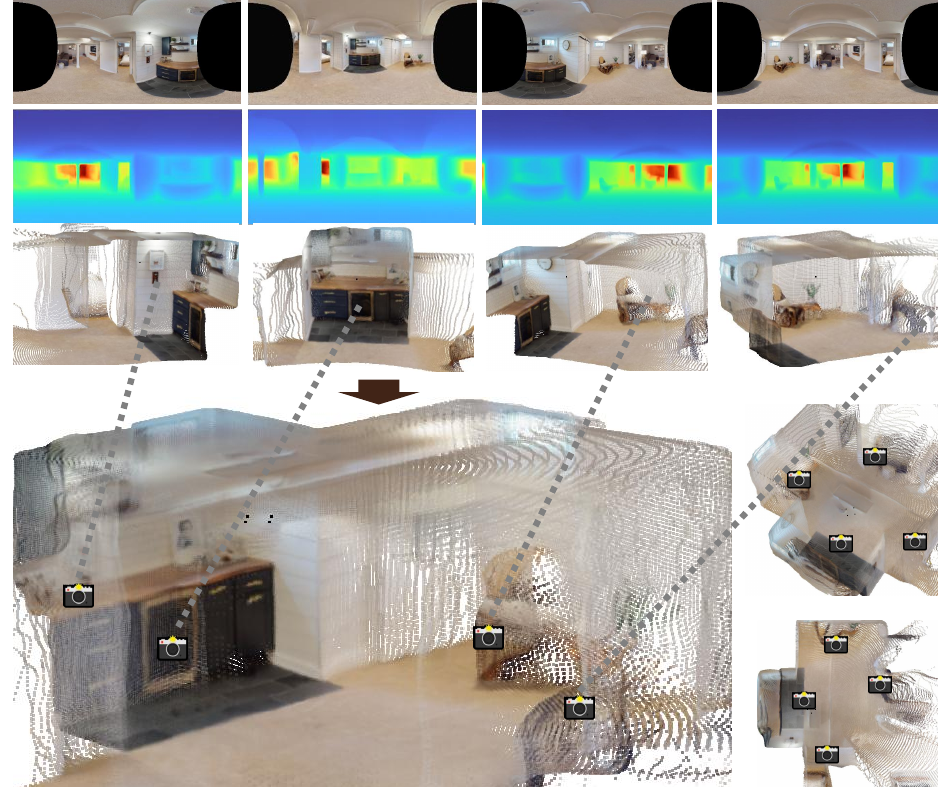}
  \caption{Real-world performance demonstration of FastViDAR. First row: ERP images converted from 220° FOV fisheye images. Second row: ERP depth maps output by FastViDAR for each viewpoint. Third row: Predicted point clouds corresponding to each viewpoint. Fourth row: Predicted fused point clouds and omnidirectional fisheye camera orientations.}
  \label{fig:homepage-demo}
\end{figure}

\textbf{Our approach.} To address the real-time inference challenge of transformer-based multi-view models like VGGT~\cite{Wang2025VGGT}, we propose FastViDAR, an efficient omnidirectional depth system that processes multiple fisheye images and produces a full $360^\circ$ depth map along with per-camera depth, fusion depth, and confidence estimates in real time, as demonstrated in Figure~\ref{fig:homepage-demo}. While our experiments demonstrate the method using four-camera setups, FastViDAR supports arbitrary numbers of cameras and arbitrary FOV cameras including ultra-wide FOV ($>180^\circ$). We project each fisheye image to a unified ERP representation, avoiding the need for the model to learn various fisheye lens distortion parameters and allowing it to focus on the ERP representation. Building on hierarchical/windowed attention~\cite{Hatamizadeh2024FasterViT}, we introduce an Alternative Hierarchical Attention (AHA) mechanism that alternates frame-local windowed self-attention with cross-frame attention over corresponding windows. Cross-view tokens are concatenated and blended via MLPs, acting as global tokens that propagate depth cues without explicit 3D or 4D cost volumes. Compared to pure window attention methods, AHA achieves good cross-frame self-attention capability with less than 10\% additional computational and memory overhead, enabling accurate depth prediction for regions outside each frame's FOV and improving multi-frame depth consistency. Compared to full attention, AHA achieves around 16× inference speed improvement in theory. In practice, FastViDAR achieves 3.3× speedup over VGGT at 640×320 resolution with 4 frames, with this advantage expanding as input resolution or frame count increases.

\textbf{Evaluation.} On real-world benchmarks including HM3D~\cite{HM3D2021} and Stanford 2D-3D-S~\cite{Armeni2017S2D3DS}, our AHA and ERP fusion methods contribute to accuracy and robustness. FastViDAR shows competitive performance compared to recent omnidirectional stereo and transformer baselines, and generalizes zero-shot to real panoramic data, delivering dense, accurate $360^\circ$ depth in real time. On embedded hardware platforms such as NVIDIA Orin NX, our method achieves up to 20 FPS inference speed with TensorRT fp16 optimization while maintaining high accuracy, demonstrating its practicality for real-world robotic applications.

\textbf{Contributions.} Our main contributions are:
\begin{enumerate}
    \item \textbf{Alternative Hierarchical Attention (AHA).} We introduce a novel attention mechanism that efficiently fuses features across views through separate intra-frame and inter-frame windowed self-attention, achieving cross-view feature mixing with reduced overhead while enabling real-time processing on embedded hardware.
    \item \textbf{ERP Fusion Method.} We propose a novel ERP fusion approach that projects multi-view depth estimates to a shared equirectangular coordinate system, enabling seamless $360^\circ$ depth fusion without expensive point cloud alignment.
    \item \textbf{ERP-aware Dataset Generation and Evaluation.} We generate ERP-aware image-depth pairs using HM3D and 2D3D-S datasets for comprehensive evaluation, demonstrating competitive zero-shot performance on real $360^\circ$ datasets while maintaining real-time processing capabilities on embedded platforms.
\end{enumerate}

\section{Related Work}
\textbf{Omnidirectional multi-view depth.}
Early deep models extend stereo to fisheye by projecting to the sphere and building cost volumes; later works improve efficiency and accuracy via spherical plane sweeping and multi-stage aggregation. Recent real-time systems rectify multi-fisheye inputs to stereo panoramas~\cite{Xie2023OmniVidar} or adopt Cassini-like projections with lightweight stereo backbones and fusion~\cite{Deng2025OmniStereo}. These achieve strong accuracy/speed but usually assume fixed calibration and rely on explicit cost volumes. In contrast, FastViDAR fuses arbitrary views with the proposed AHA without constructing stereo volumes and employs ERP fusion for seamless $360^\circ$ depth estimation, while flexibly outputting metric depth for each individual view.

\textbf{Panoramic/fisheye monocular depth.}
Standard pinhole networks face challenges with limited field of view and multi-view depth consistency. Specialized spherical/cubemap representations mitigate distortion, while DAC~\cite{DAC2025} attains zero-shot metric depth by mapping any input to a standard ERP representation. However, monocular methods still face scale ambiguity and cannot exploit cross-view constraints. FastViDAR also uses ERP representation but employs global attention across frames to achieve implicit multi-view scale constraints and better scale consistency.

\textbf{Efficient stereo and MVS.}
Efficiency-oriented designs reduce the burden of 3D cost volumes via 2D aggregation and lightweight modules~\cite{LightStereo2024}. Coarse-to-fine multi-view stereo (e.g., MVSNet/CasMVSNet~\cite{Yao2018MVSNet, Yao2019CasMVSNet}) limits depth hypotheses. CasOmniMVS~\cite{CasOmniMVS2024} adapts spherical sweeping density for omnidirectional scenes. Our method avoids explicit volumes altogether and relies on learned implicit correlations via AHA.

\textbf{Transformers for 3D perception.}
Alternating local/global attention has proven effective for multi-view geometry and camera prediction~\cite{Wang2025VGGT}. Hierarchical/windowed attention such as FasterViT~\cite{Hatamizadeh2024FasterViT} scales attention with carrier tokens. FastViDAR tailors this paradigm to multi-camera omnidirectional depth and employs ERP fusion for improved geometric consistency across views.

\section{Method}

\subsection{Fisheye Camera and Equirectangular Projection (ERP)}
\label{sec:erp}
\begin{figure}[t]
  \centering
  \includegraphics[width=0.95\linewidth,trim={20 180 105 0},clip]{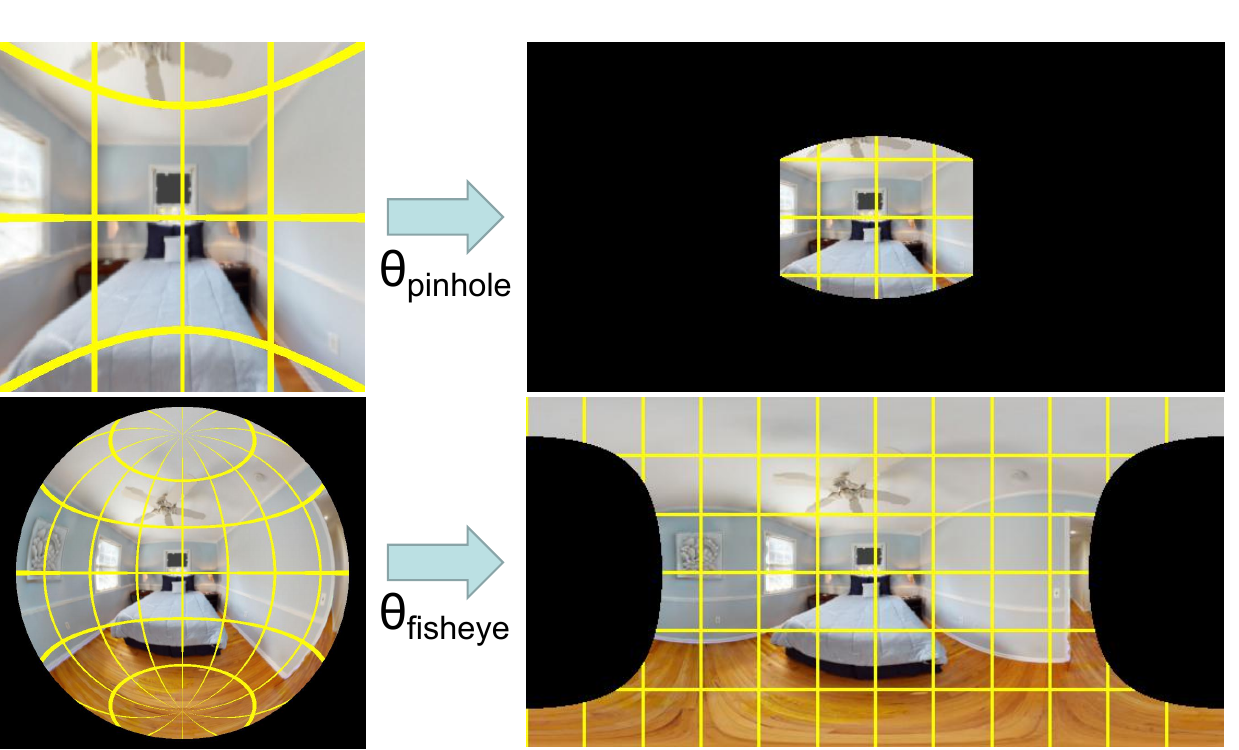}
  \vspace{0.1em}
  \caption{Unified ERP representation for different camera types. Top row: pinhole camera (FOV=$100^\circ$) with intrinsics $\theta_{\text{pinhole}}$ projects to ERP. Bottom row: fisheye camera (FOV=$220^\circ$) with intrinsics $\theta_{\text{fisheye}}$ projects to the same ERP lattice. Yellow lines show the correspondence between ERP lattice positions and the original camera-specific lattice positions.}
  \label{fig:erp_roundtrip}
\end{figure}

We adopt \emph{equirectangular projection} (ERP) images as the network input to decouple lens-specific intrinsics from learning. Any \emph{central} fisheye model (e.g., KB/equidistant/equisolid~\cite{Kannala2006Generic}, OCamCalib polynomial~\cite{Scaramuzza2006OCamCalib}, Double Sphere (DSCamera)~\cite{Usenko2018DoubleSphere}, unified central~\cite{Geyer2000Unified}) maps a pixel $\mathbf{u}=(u,v)$ to a unit viewing ray $\mathbf{d}=(d_x,d_y,d_z)\in\mathbb{S}^2$ in the camera frame via $\pi^{-1}_{\theta}$ (intrinsics $\theta$). After this step, ERP coordinates are a \emph{pure spherical reparameterization} independent of $\theta$; we take $+z$ forward, $+x$ right, $+y$ up, ERP size $W{\times}H$ with origin at top-left, and longitude/latitude $(\lambda,\phi)$ in radians:
\begin{align}
\lambda&=\operatorname{atan2}(d_x,d_z), & \phi&=\arcsin(d_y),\label{eq:erp-coords}\\
x&=\Big(\tfrac{\lambda}{2\pi}+\tfrac{1}{2}\Big)W, &
y&=\Big(\tfrac{1}{2}-\tfrac{\phi}{\pi}\Big)H.\label{eq:erp-pixels}
\end{align}
Thus images from heterogeneous fisheye lenses land on the same ERP lattice, as shown in Figure~\ref{fig:erp_roundtrip}, allowing the network to learn on a stable, camera-agnostic domain while preserving the native wide FOV that perspective pinhole would crop away. Although ERP introduces polar area distortion (local scale $\propto 1/\cos\phi$), our experiments show the proposed method handles it well. For visualization or synthesis, the reverse path is direct: from ERP $(x,y)$ recover $(\lambda,\phi)$, form $\mathbf{d}=[\sin\lambda\cos\phi,\,\sin\phi,\,\cos\lambda\cos\phi]^\top$, and project via $\mathbf{u}=\pi_{\theta}(\mathbf{d})$ (e.g., DSCamera) to render fisheye views within the lens FOV.

\begin{figure*}[t]
        \centering
        \includegraphics[width=\textwidth,trim={5 290 270 80},clip]{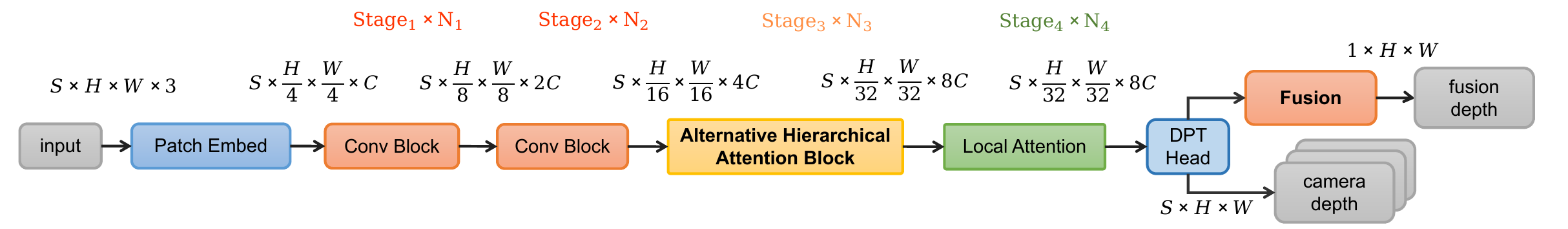}
        \vspace{-1.2em}
        \caption{FastViDAR architecture overview. Stages~1-2: convolutional stem for pyramidal feature extraction. Stage~3: AHA blocks that alternate window self-attention on \emph{local tokens} with frame- and global-level self-attention on pooled \emph{summary tokens}. Stage~4: local refinement via stacked window self-attention.}
      
        \label{fig:overview}
        \vspace{-0.8em}
      \end{figure*}

\subsection{Alternative Hierarchical Attention (AHA)}
\label{sec:aha}

Monocular depth is fundamentally \emph{scale ambiguous}. Under a calibrated pinhole model, if 3D points $\mathbf{X}\!\in\!\mathbb{R}^3$ and camera pose $(\mathbf{R},\mathbf{t})$ explain the observations, then for any $\alpha>0$, the projected image coordinates satisfy
\begin{equation}
\mathbf{x}\ \propto\ \mathbf{K}\!\left(\mathbf{R}(\alpha\mathbf{X})+\alpha\mathbf{t}\right)
\ =\ \alpha\,\mathbf{K}\!\left(\mathbf{R}\mathbf{X}+\mathbf{t}\right)\ \propto\ \mathbf{K}\!\left(\mathbf{R}\mathbf{X}+\mathbf{t}\right)\label{eq:scale-ambiguity}
\end{equation}
where $\mathbf{x}=(u,v,1)^\top$ represents homogeneous image coordinates and $\mathbf{K}$ is the camera intrinsic matrix. Thus, depths are identifiable only up to a global scale in the absence of absolute metric cues (e.g., known baseline, object size priors, or other metric information). Multi-frame modeling constrains \emph{relative} structure and motion; however, na\"{i}vely applying global self-attention over all tokens from $S$ frames with $N$ tokens each incurs quadratic cost $\mathcal{O}\!\left((S\,N)^2\right)$ in time and memory. VGGT~\cite{Wang2025VGGT} leverages multi-frame attention effectively yet remains computationally heavy for embedded deployment.
Inspired by FasterViT~\cite{Hatamizadeh2024FasterViT} and convolutional tokenization as in CvT~\cite{wu_cvt_2021}, 
we propose \emph{Alternative Hierarchical Attention} (AHA), which applies window self-attention on \emph{local tokens}
and alternates frame- and global-level self-attention on compact \emph{summary tokens}.

\begin{figure*}[t]
  \centering
  \includegraphics[width=\textwidth,trim={30 280 20 30},clip]{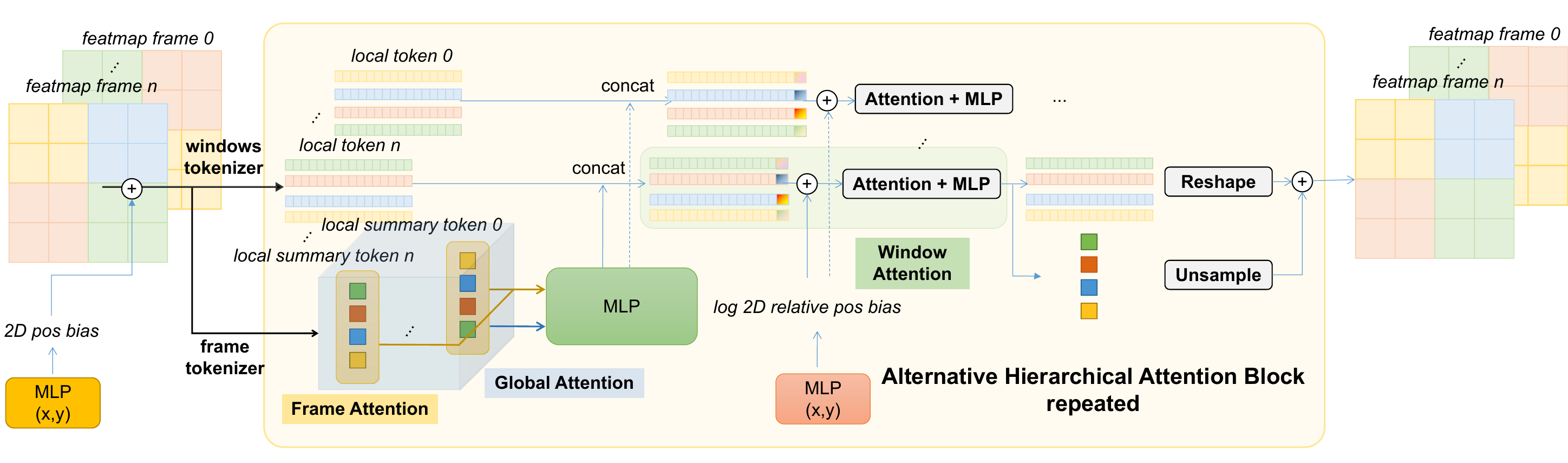}
  \vspace{-1.2em}
  \caption{AHA block. Window self-attention over local tokens, frame-level self-attention over summary tokens and
  global self-attention over all summaries.}
  \label{fig:aha-block}
  \vspace{-0.8em}
\end{figure*}

\paragraph{Overview and backbone}
As shown in Figure~\ref{fig:overview}, the model comprises \emph{four stacked stages} and supports an arbitrary number of frames $S$ (e.g., 4-camera rigs or temporal clips) and arbitrary input resolutions, enabled by adaptive padding and average pooling. While our experiments use $S=4$ cameras, the method is designed to handle arbitrary numbers of cameras. The method is pose-flexible, accommodating arbitrary camera configurations and orientations without requiring specific geometric constraints.
Stages~1-2 form a convolutional stem composed of repeated
\texttt{conv2d}-\texttt{bn}-\texttt{gelu} 
 blocks with downsampling at stage boundaries, producing feature maps
$F \in \mathbb{R}^{B \times S \times C \times H \times W}$,
where $B$ is the batch size, $S$ the number of frames/cameras, $C$ channels, and $H{\times}W$ the input spatial size. Stage~$i$ outputs resolution $H/2^{i+2} \times W/2^{i+2}$ with $\approx 2^i C$ channels.
Stage~3 stacks $L$ \emph{AHA blocks} that apply window self-attention on \emph{local tokens} and alternate frame- and global-level self-attention on compact \emph{summary tokens}, as detailed in Figure~\ref{fig:aha-block}.
Stage~4 performs \emph{local refinement} via a stack of self-attention layers applied to the windowed local tokens.

\paragraph{Tokenizers and notation}
We form two token sets from the stage-2 feature map $F\!\in\!\mathbb{R}^{B\times S\times C\times H\times W}$ by applying learnable positional bias to the feature maps.
Windows are \emph{non-overlapping} with size $P_h\times P_w$ and stride $(P_h,P_w)$.
To ensure complete window coverage, we use adaptive padding to $H'=\lceil H/P_h\rceil P_h$ and $W'=\lceil W/P_w\rceil P_w$, then partition each frame into $N_h=H'/P_h$ by $N_w=W'/P_w$ windows; the number of windows per frame is $M=N_hN_w$.
We index windows within a frame by $m\in\{1,\ldots,M\}$ (row-major over the $N_h\times N_w$ grid), and positions inside a window by $p\in\{1,\ldots,P_hP_w\}$.
For clarity, we denote the per-frame \emph{local tokens} by $L_{b,s}\!\in\!\mathbb{R}^{M\times (P_hP_w)\times C}$ and the \emph{local summary tokens} by $S^{\text{loc}}_{b,s}\!\in\!\mathbb{R}^{M\times C}$.

\noindent\textbf{Tokenizers as operators (input $\to$ output shape):}
\begin{align}
\mathrm{WinTok}_{(P_h,P_w)}:\;\mathbb{R}^{C\times H'\times W'} &\to \mathbb{R}^{M\times (P_hP_w)\times C},\label{eq:win-tok}\\
\mathrm{FrameTok}:\;\mathbb{R}^{M\times (P_hP_w)\times C} &\to \mathbb{R}^{M\times C}.\label{eq:frame-tok}
\end{align}

\noindent
(i) \emph{Local tokens.} For a padded frame feature map $F_{b,s}\!\in\!\mathbb{R}^{C\times H'\times W'}$, we define
\[
L_{b,s} \;\coloneqq\; \mathrm{WinTok}_{(P_h,P_w)}\!\big(F_{b,s}\big)\;\in\;\mathbb{R}^{M\times (P_hP_w)\times C}.
\]

\noindent
(ii) \emph{Local summary tokens.} Per-window pooled descriptors are given by
\[
S^{\text{loc}}_{b,s} \;\coloneqq\; \mathrm{FrameTok}\!\big(L_{b,s}\big)\;\in\;\mathbb{R}^{M\times C}.
\]
Elementwise, for window index $m$, we perform average pooling over the window dimension:
\begin{equation}
S^{\text{loc}}_{b,s}[m] \;=\; \tfrac{1}{P_hP_w}\!\sum_{p=1}^{P_hP_w}\! L_{b,s}[m,p,:]. \label{eq:summary-aniso}
\end{equation}

\noindent
Stacking batch and frames yields
$L\in\mathbb{R}^{(BS)M\times (P_hP_w)\times C}$ and
$S^{\text{loc}}\in\mathbb{R}^{(BS)\times M\times C}$.

\vspace{0.25em}
\paragraph{Three-level attention in AHA}
Each AHA block alternates three attentions, as shown in Figure~\ref{fig:aha-block}. 
Here, self-attention refers to Multi-Head Self-Attention (MHSA)~\cite{Vaswani2017Attention}:

\begin{enumerate}
\item \textbf{Window attention (local).}
Self-attention is applied \emph{within} each window independently.
For $m\!\in\!\{1,\ldots,M\}$,
\begin{align}
\mathrm{Attn}_{\text{win}}:\; &L_{b,s}[m]\in\mathbb{R}^{(P_hP_w)\times C}\nonumber\\
&\;\mapsto\; \tilde{L}_{b,s}[m]\in\mathbb{R}^{(P_hP_w)\times C},\label{eq:win-attn}
\end{align}
with relative positional bias inside the window.

\item \textbf{Frame attention (per-frame summaries).}
Self-attention over summary tokens \emph{within} a single frame:
\begin{align}
\mathrm{Attn}_{\text{frame}}:\; &S^{\text{loc}}_{b,s}\in\mathbb{R}^{M\times C}\nonumber\\
&\;\mapsto\; \hat{S}_{b,s}\in\mathbb{R}^{M\times C}.\label{eq:frame-attn}
\end{align}
A learnable frame/camera embedding $\mathbf{e}_s$ is added to encode view/time identity.

\item \textbf{Global attention (multi-frame summaries).}
Self-attention over all summary tokens across $S$ frames:
\begin{align}
\mathrm{Attn}_{\text{global}}:\; &\mathrm{concat}_s\!\big(\hat{S}_{b,s}\big)\in\mathbb{R}^{(SM)\times C}\nonumber\\
&\;\mapsto\; \bar{S}_{b}\in\mathbb{R}^{(SM)\times C}.\label{eq:global-attn}
\end{align}
This fuses cross-view or temporal context that helps resolve scale ambiguity and improves geometric consistency.
\end{enumerate}

\paragraph{Local refinement via stacked window self-attention}
Stage~4 is a \emph{purely local} refinement: we apply $N_4$ layers of window MHSA on the same $(P_h,P_w)$ 
partition.
For each $(b,s,m)$, let $X^{(0)}\!=\!L_{b,s}[m]\in\mathbb{R}^{(P_hP_w)\times C}$.
Each layer uses pre-norm MHSA with a 1-layer MLP:
\begin{align}
X^{(r)} &= X^{(r-1)} + \mathrm{MHSA}_{\text{win}}\!\big(\mathrm{LN}(X^{(r-1)});\,B_{\mathrm{rel}}\big),\nonumber\\
X^{(r)} &\leftarrow X^{(r)} + \mathrm{MLP}\!\big(\mathrm{LN}(X^{(r)})\big),\quad r=1,\ldots,R,\label{eq:local-refinement}
\end{align}
where $B_{\mathrm{rel}}$ is the within-window relative positional bias.
Stacking all windows yields $L^\star_{b,s}\!\in\!\mathbb{R}^{M\times(P_hP_w)\times C}$.

\paragraph{Complexity}
Let $N\!=\!H'W'$ be the number of dense tokens per frame (after padding), $S$ the number of frames, and
$P\!=\!P_hP_w$ the window size in tokens (so $M\!=\!N/P$ windows per frame).
We report leading-order costs \emph{per head} and the size of the attention matrix (which dominates activation memory), ignoring linear projections/MLPs.

\noindent\textbf{VGGT (full attention over all dense tokens).}
Sequence length $L_{\text{full}}=SN$:
\begin{equation}
\text{Compute}=\mathcal{O}\big((SN)^2\big),\qquad
\text{Memory}=\mathcal{O}\big((SN)^2\big).\label{eq:vggt-complexity}
\end{equation}

\noindent\textbf{AHA (ours).}
Sequence lengths: window $P$, per-frame summaries $M=N/P$, global summaries $SM=SN/P$.
\begin{align}
&\underbrace{\mathcal{O}\big(S\,M\,P^2\big)}_{\text{window attention}}
\;+\;
\underbrace{\mathcal{O}\big(S\,M^2\big)}_{\text{frame attention}}
\;+\;
\underbrace{\mathcal{O}\big((SM)^2\big)}_{\text{global attention}}\nonumber\\
&\;=\;
\mathcal{O}\big(SNP\;+\;(SN/P)^2\big)\label{eq:aha-complexity}
\end{align}
with the same orders for activation memory (attention matrices of sizes $SNP$, $S\,M^2$, and $(SM)^2$, respectively).

\noindent\textbf{Comparison (per block, per head).}
\begin{align}
\frac{\mathrm{AHA}}{\mathrm{Full}}
&= \frac{SNP + (SN/P)^2}{(SN)^2}
= \frac{P}{SN} + \frac{1}{P^2},\label{eq:complexity-ratio}
\end{align}
where $S$ is the number of frames, $N$ is the tokens per frame after downsampling, and $P{=}W_h W_w$ is the number of tokens in a window (e.g., $7{\times}7 \Rightarrow P{=}49$).
For an input of $640{\times}320$ pixels with $7{\times}7$ windows, we have $N \!\approx\! 200$ and with $S{=}4$ the ratio becomes
$\frac{\mathrm{AHA}}{\mathrm{Full}} \approx \frac{49}{4\cdot 200}+\frac{1}{49^2} \approx 0.0604$,
i.e., a $\sim\!16{\times}$ reduction.

As $SN$ grows, the linear term $\tfrac{P}{SN}$ vanishes and the \emph{ratio} saturates at
$\lim_{SN\to\infty}\tfrac{\mathrm{AHA}}{\mathrm{Full}}=\tfrac{1}{P^2}$,
so the maximal theoretical speedup is $P^2$ (for $P{=}49$, $\sim\!2401{\times}$).
Equivalently, when $SN \gg P^3$, AHA's \emph{absolute} cost is dominated by its grouped quadratic term $(SN/P)^2$, but the \emph{relative} complexity no longer decreases and remains near $1/P^2$.

\begin{figure*}[t]
  \centering
  \vspace{1em}
  
  \renewcommand{\arraystretch}{0.05}
  \setlength{\tabcolsep}{0pt}
  \setlength{\arrayrulewidth}{0pt}
  \begin{tabular}{@{}c@{\hspace{-6pt}}c@{\hspace{-6pt}}c@{\hspace{-6pt}}c@{\hspace{-6pt}}c@{\hspace{-6pt}}c@{}}
    & \tiny\textbf{Camera 1} & \tiny\textbf{Camera 2} & \tiny\textbf{Camera 3} & \tiny\textbf{Camera 4} & \tiny\textbf{Fusion} \\
    \\[1pt]
    \multicolumn{1}{c|}{\rotatebox{90}{\tiny\textbf{\ \ \ \ Input Images}}} &
    \includegraphics[width=0.194\textwidth]{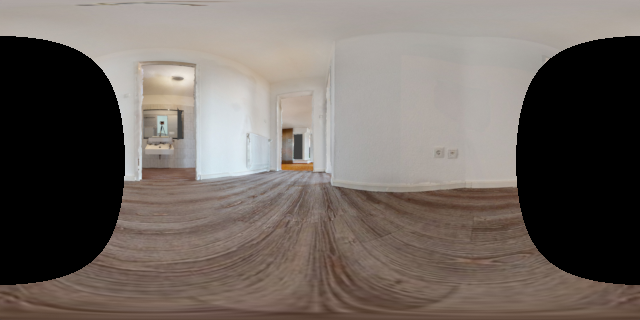} &
    \includegraphics[width=0.194\textwidth]{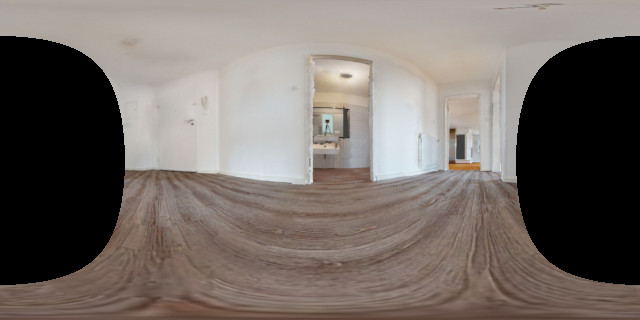} &
    \includegraphics[width=0.194\textwidth]{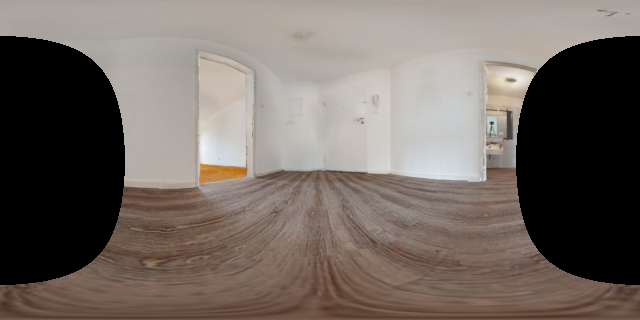} &
    \includegraphics[width=0.194\textwidth]{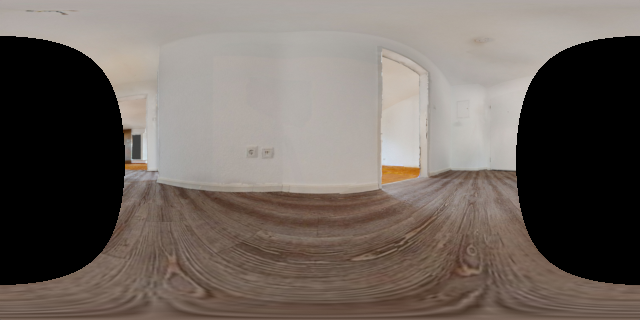} &
    \includegraphics[width=0.194\textwidth]{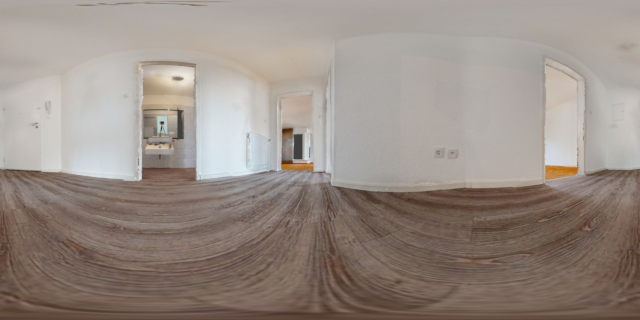} \\
    \\[-3pt]
    
    \multicolumn{1}{c|}{\rotatebox{90}{\tiny\textbf{\ \ \ \ Ground Truth}}} &
    \begin{tikzpicture}
      \node[anchor=center] (img1) {\includegraphics[width=0.194\textwidth]{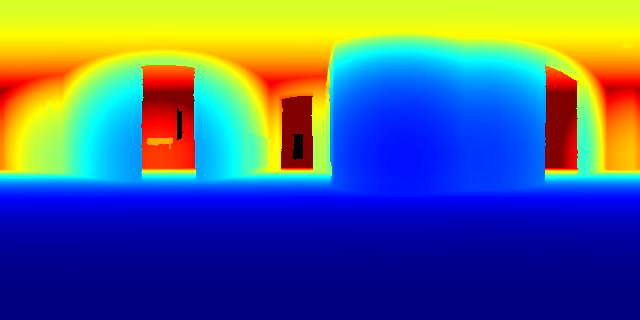}};
      \draw[red, dashed, thick] (img1.south west) ++(0.05*0.2*\textwidth, 0.25*0.195*0.6*\textwidth) rectangle ++(0.125*0.2*\textwidth, 0.5*0.195*0.6*\textwidth);
      \draw[red, dashed, thick] (img1.south west) ++(0.9*0.2*\textwidth, 0.25*0.195*0.6*\textwidth) rectangle ++(0.125*0.2*\textwidth, 0.5*0.195*0.6*\textwidth);
    \end{tikzpicture} &
    \begin{tikzpicture}
      \node[anchor=center] (img2) {\includegraphics[width=0.194\textwidth]{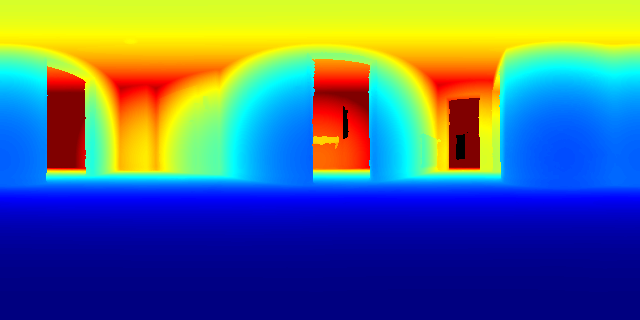}};
      \draw[red, dashed, thick] (img2.south west) ++(0.05*0.2*\textwidth, 0.25*0.195*0.6*\textwidth) rectangle ++(0.125*0.2*\textwidth, 0.5*0.195*0.6*\textwidth);
      \draw[red, dashed, thick] (img2.south west) ++(0.9*0.2*\textwidth, 0.25*0.195*0.6*\textwidth) rectangle ++(0.125*0.2*\textwidth, 0.5*0.195*0.6*\textwidth);
    \end{tikzpicture} &
    \begin{tikzpicture}
      \node[anchor=center] (img3) {\includegraphics[width=0.194\textwidth]{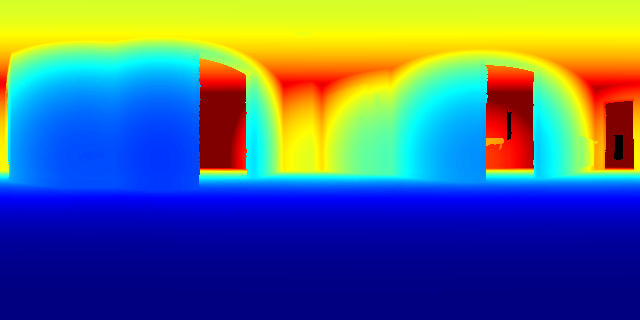}};
      \draw[red, dashed, thick] (img3.south west) ++(0.05*0.2*\textwidth, 0.25*0.195*0.6*\textwidth) rectangle ++(0.125*0.2*\textwidth, 0.5*0.195*0.6*\textwidth);
      \draw[red, dashed, thick] (img3.south west) ++(0.9*0.2*\textwidth, 0.25*0.195*0.6*\textwidth) rectangle ++(0.125*0.2*\textwidth, 0.5*0.195*0.6*\textwidth);
    \end{tikzpicture} &
    \begin{tikzpicture}
      \node[anchor=center] (img4) {\includegraphics[width=0.194\textwidth]{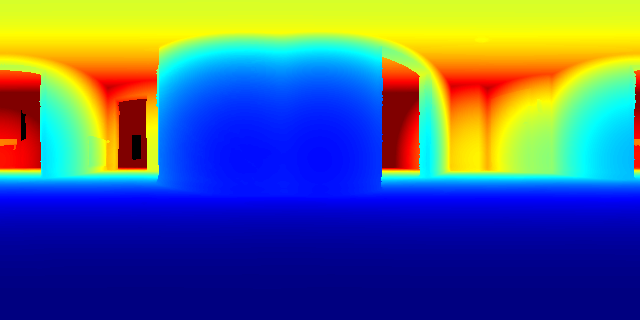}};
      \draw[red, dashed, thick] (img4.south west) ++(0.05*0.2*\textwidth, 0.25*0.195*0.6*\textwidth) rectangle ++(0.125*0.2*\textwidth, 0.5*0.195*0.6*\textwidth);
      \draw[red, dashed, thick] (img4.south west) ++(0.9*0.2*\textwidth, 0.25*0.195*0.6*\textwidth) rectangle ++(0.125*0.2*\textwidth, 0.5*0.195*0.6*\textwidth);
    \end{tikzpicture} &
    \begin{tikzpicture}
      \node[anchor=center] (img5) {\includegraphics[width=0.194\textwidth]{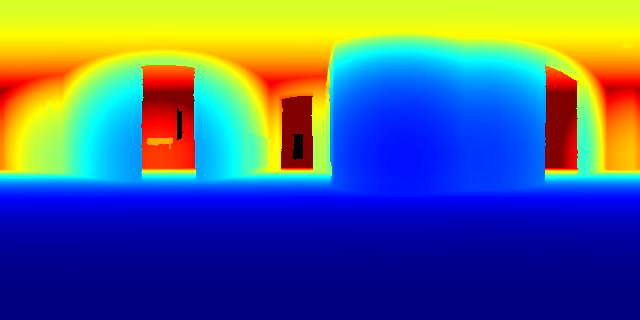}};
    \end{tikzpicture} \\[-3pt]
    
    \multicolumn{1}{c|}{\rotatebox{90}{\tiny\textbf{\ \ \ \ \ \ No-Global }}} &
    \begin{tikzpicture}
      \node[anchor=center] (img1) {\includegraphics[width=0.194\textwidth]{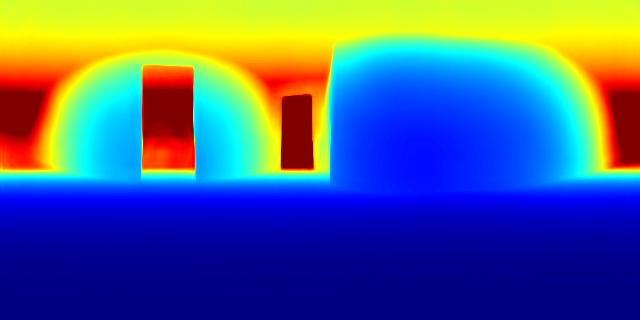}};
      \draw[red, dashed, thick] (img1.south west) ++(0.05*0.2*\textwidth, 0.25*0.195*0.6*\textwidth) rectangle ++(0.125*0.2*\textwidth, 0.5*0.195*0.6*\textwidth);
      \draw[red, dashed, thick] (img1.south west) ++(0.9*0.2*\textwidth, 0.25*0.195*0.6*\textwidth) rectangle ++(0.125*0.2*\textwidth, 0.5*0.195*0.6*\textwidth);
    \end{tikzpicture} &
    \begin{tikzpicture}
      \node[anchor=center] (img2) {\includegraphics[width=0.194\textwidth]{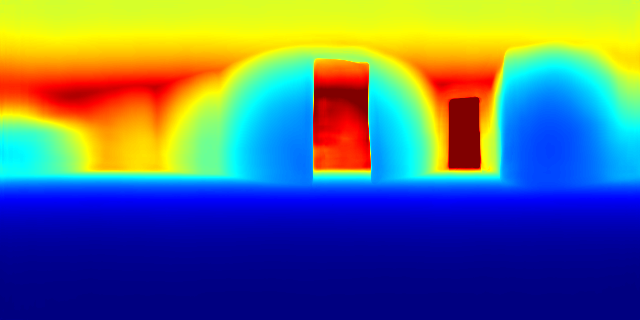}};
      \draw[red, dashed, thick] (img2.south west) ++(0.05*0.2*\textwidth, 0.25*0.195*0.6*\textwidth) rectangle ++(0.125*0.2*\textwidth, 0.5*0.195*0.6*\textwidth);
      \draw[red, dashed, thick] (img2.south west) ++(0.9*0.2*\textwidth, 0.25*0.195*0.6*\textwidth) rectangle ++(0.125*0.2*\textwidth, 0.5*0.195*0.6*\textwidth);
    \end{tikzpicture} &
    \begin{tikzpicture}
      \node[anchor=center] (img3) {\includegraphics[width=0.194\textwidth]{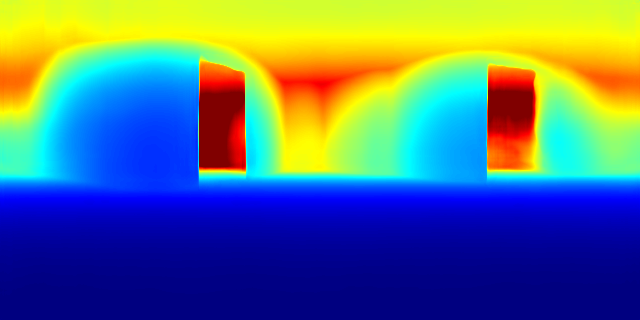}};
      \draw[red, dashed, thick] (img3.south west) ++(0.05*0.2*\textwidth, 0.25*0.195*0.6*\textwidth) rectangle ++(0.125*0.2*\textwidth, 0.5*0.195*0.6*\textwidth);
      \draw[red, dashed, thick] (img3.south west) ++(0.9*0.2*\textwidth, 0.25*0.195*0.6*\textwidth) rectangle ++(0.125*0.2*\textwidth, 0.5*0.195*0.6*\textwidth);
    \end{tikzpicture} &
    \begin{tikzpicture}
      \node[anchor=center] (img4) {\includegraphics[width=0.194\textwidth]{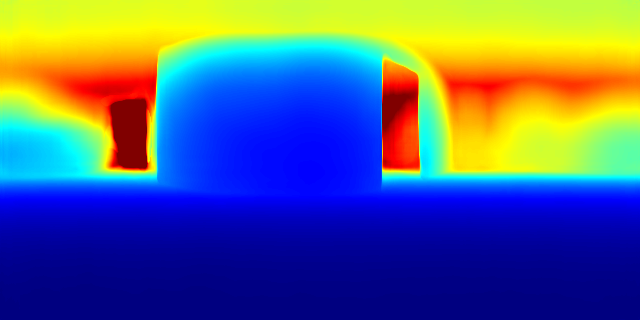}};
      \draw[red, dashed, thick] (img4.south west) ++(0.05*0.2*\textwidth, 0.25*0.195*0.6*\textwidth) rectangle ++(0.125*0.2*\textwidth, 0.5*0.195*0.6*\textwidth);
      \draw[red, dashed, thick] (img4.south west) ++(0.9*0.2*\textwidth, 0.25*0.195*0.6*\textwidth) rectangle ++(0.125*0.2*\textwidth, 0.5*0.195*0.6*\textwidth);
    \end{tikzpicture} &
    \begin{tikzpicture}
      \node[anchor=center] (img5) {\includegraphics[width=0.194\textwidth]{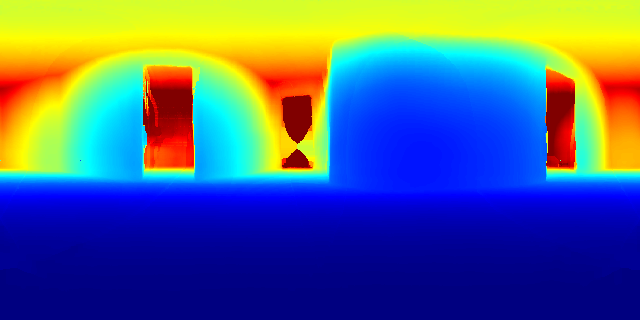}};
    \end{tikzpicture} \\[-3pt]
    
    \multicolumn{1}{c|}{\rotatebox{90}{\tiny\textbf{\ \ \ \ \ \ Error}}} &
    \includegraphics[width=0.194\textwidth]{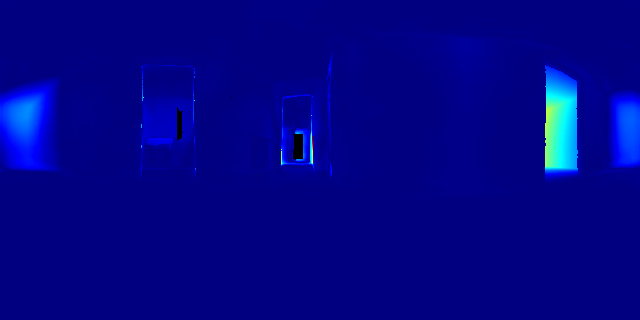} &
    \includegraphics[width=0.194\textwidth]{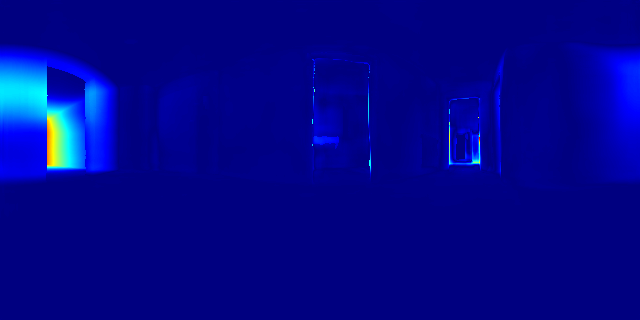} &
    \includegraphics[width=0.194\textwidth]{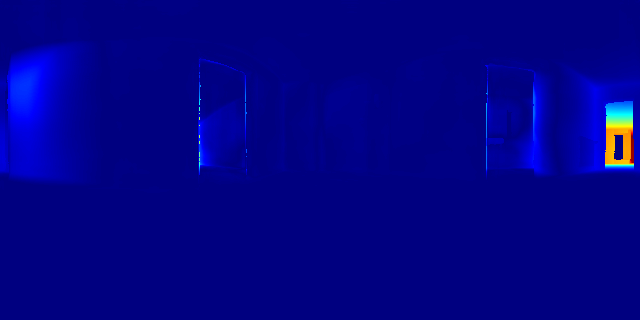} &
    \includegraphics[width=0.194\textwidth]{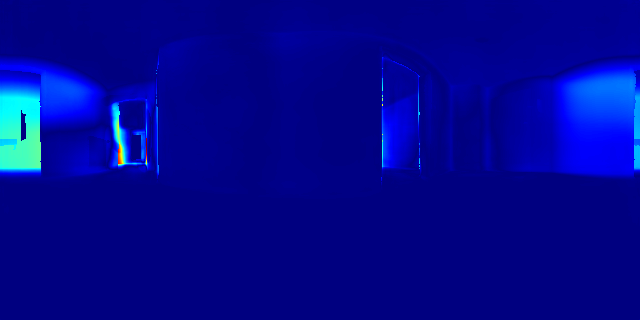} &
    \includegraphics[width=0.194\textwidth]{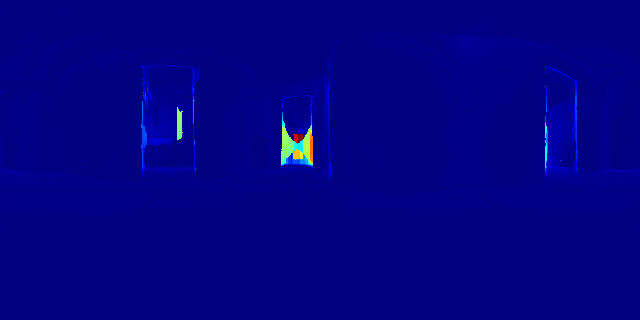} \\[-3pt]
    
    \multicolumn{1}{c|}{\rotatebox{90}{\tiny\textbf{\ \ \ \ \ \ AHA}}} &
    \begin{tikzpicture}
      \node[anchor=center] (img1) {\includegraphics[width=0.194\textwidth]{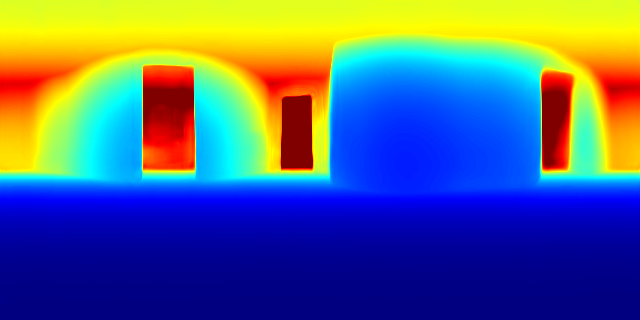}};
      \draw[red, dashed, thick] (img1.south west) ++(0.05*0.2*\textwidth, 0.25*0.195*0.6*\textwidth) rectangle ++(0.125*0.2*\textwidth, 0.5*0.195*0.6*\textwidth);
      \draw[red, dashed, thick] (img1.south west) ++(0.9*0.2*\textwidth, 0.25*0.195*0.6*\textwidth) rectangle ++(0.125*0.2*\textwidth, 0.5*0.195*0.6*\textwidth);
    \end{tikzpicture} &
    \begin{tikzpicture}
      \node[anchor=center] (img2) {\includegraphics[width=0.194\textwidth]{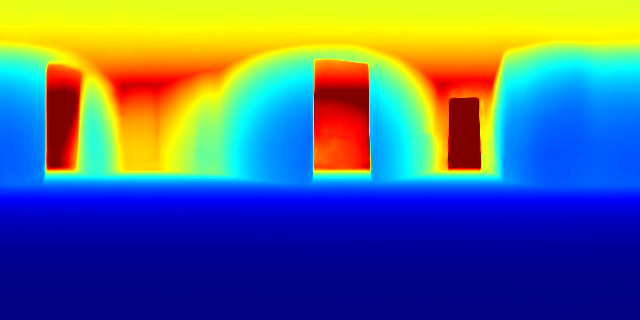}};
      \draw[red, dashed, thick] (img2.south west) ++(0.05*0.2*\textwidth, 0.25*0.195*0.6*\textwidth) rectangle ++(0.125*0.2*\textwidth, 0.5*0.195*0.6*\textwidth);
      \draw[red, dashed, thick] (img2.south west) ++(0.9*0.2*\textwidth, 0.25*0.195*0.6*\textwidth) rectangle ++(0.125*0.2*\textwidth, 0.5*0.195*0.6*\textwidth);
    \end{tikzpicture} &
    \begin{tikzpicture}
      \node[anchor=center] (img3) {\includegraphics[width=0.194\textwidth]{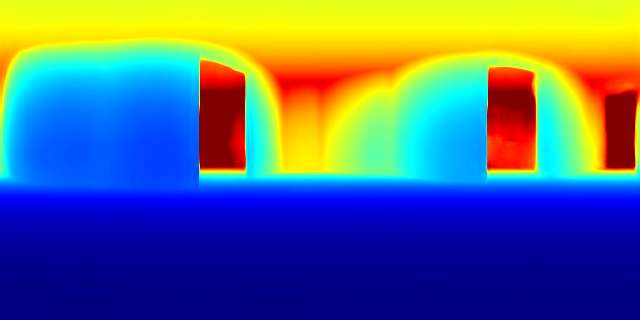}};
      \draw[red, dashed, thick] (img3.south west) ++(0.05*0.2*\textwidth, 0.25*0.195*0.6*\textwidth) rectangle ++(0.125*0.2*\textwidth, 0.5*0.195*0.6*\textwidth);
      \draw[red, dashed, thick] (img3.south west) ++(0.9*0.2*\textwidth, 0.25*0.195*0.6*\textwidth) rectangle ++(0.125*0.2*\textwidth, 0.5*0.195*0.6*\textwidth);
    \end{tikzpicture} &
    \begin{tikzpicture}
      \node[anchor=center] (img4) {\includegraphics[width=0.194\textwidth]{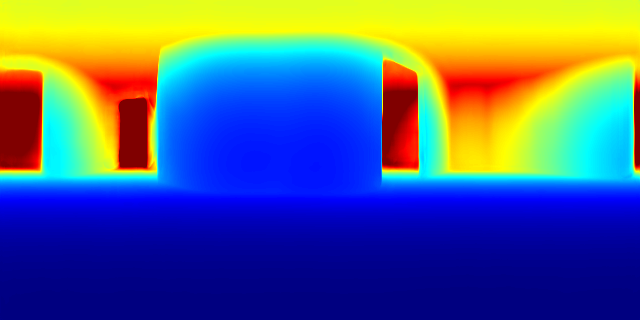}};
      \draw[red, dashed, thick] (img4.south west) ++(0.05*0.2*\textwidth, 0.25*0.195*0.6*\textwidth) rectangle ++(0.125*0.2*\textwidth, 0.5*0.195*0.6*\textwidth);
      \draw[red, dashed, thick] (img4.south west) ++(0.9*0.2*\textwidth, 0.25*0.195*0.6*\textwidth) rectangle ++(0.125*0.2*\textwidth, 0.5*0.195*0.6*\textwidth);
    \end{tikzpicture} &
    \begin{tikzpicture}
      \node[anchor=center] (img5) {\includegraphics[width=0.194\textwidth]{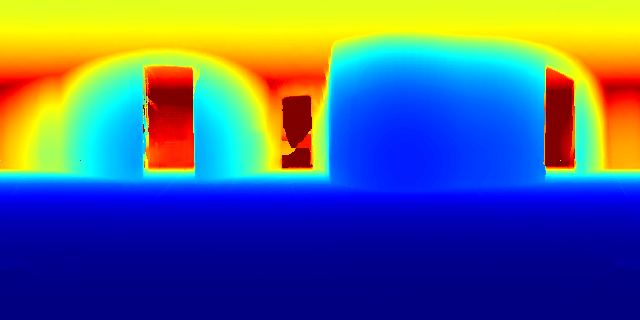}};
    \end{tikzpicture} \\[-3pt]
    
    \multicolumn{1}{c|}{\rotatebox{90}{\tiny\textbf{\ \ \ \ \ \ Error}}} &
    \includegraphics[width=0.194\textwidth]{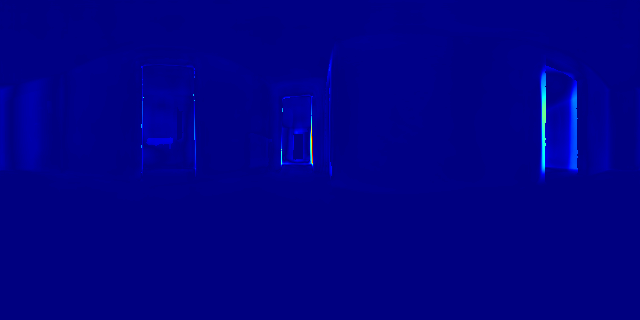} &
    \includegraphics[width=0.194\textwidth]{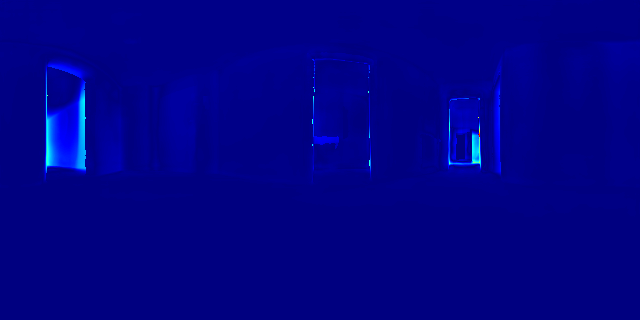} &
    \includegraphics[width=0.194\textwidth]{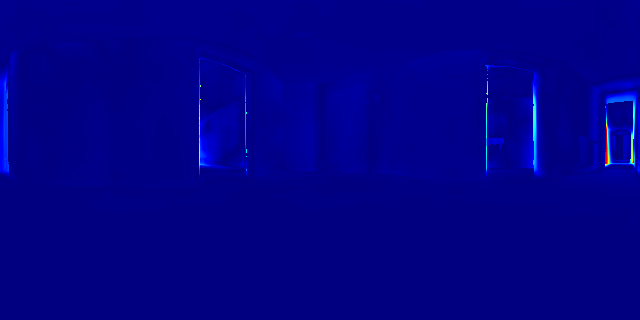} &
    \includegraphics[width=0.194\textwidth]{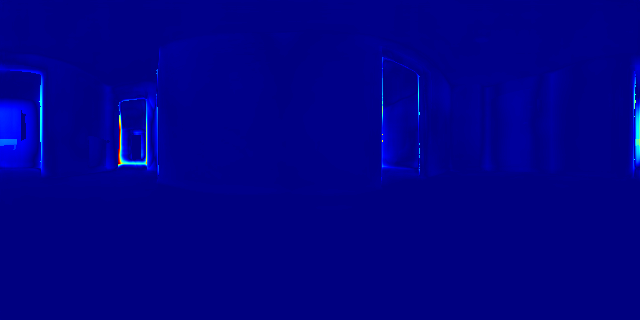} &
    \includegraphics[width=0.194\textwidth]{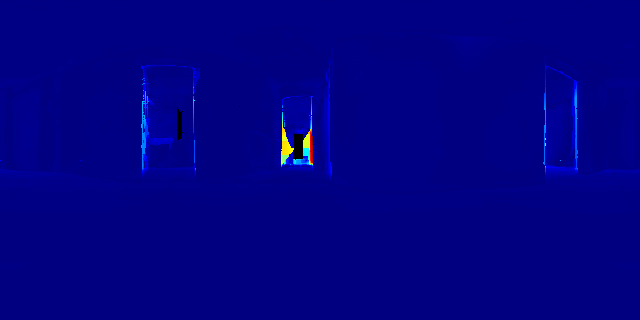} \\
  \end{tabular}
  
  \caption{Qualitative visualization of AHA vs No-Global attention comparison. The table demonstrates the performance difference between AHA (with global attention) and No-Global attention across different camera views and their fusion results. Main differences are highlighted with red dashed rectangles.}
  \label{fig:abl-visual}
  \vspace{-1.5em}
  \end{figure*}
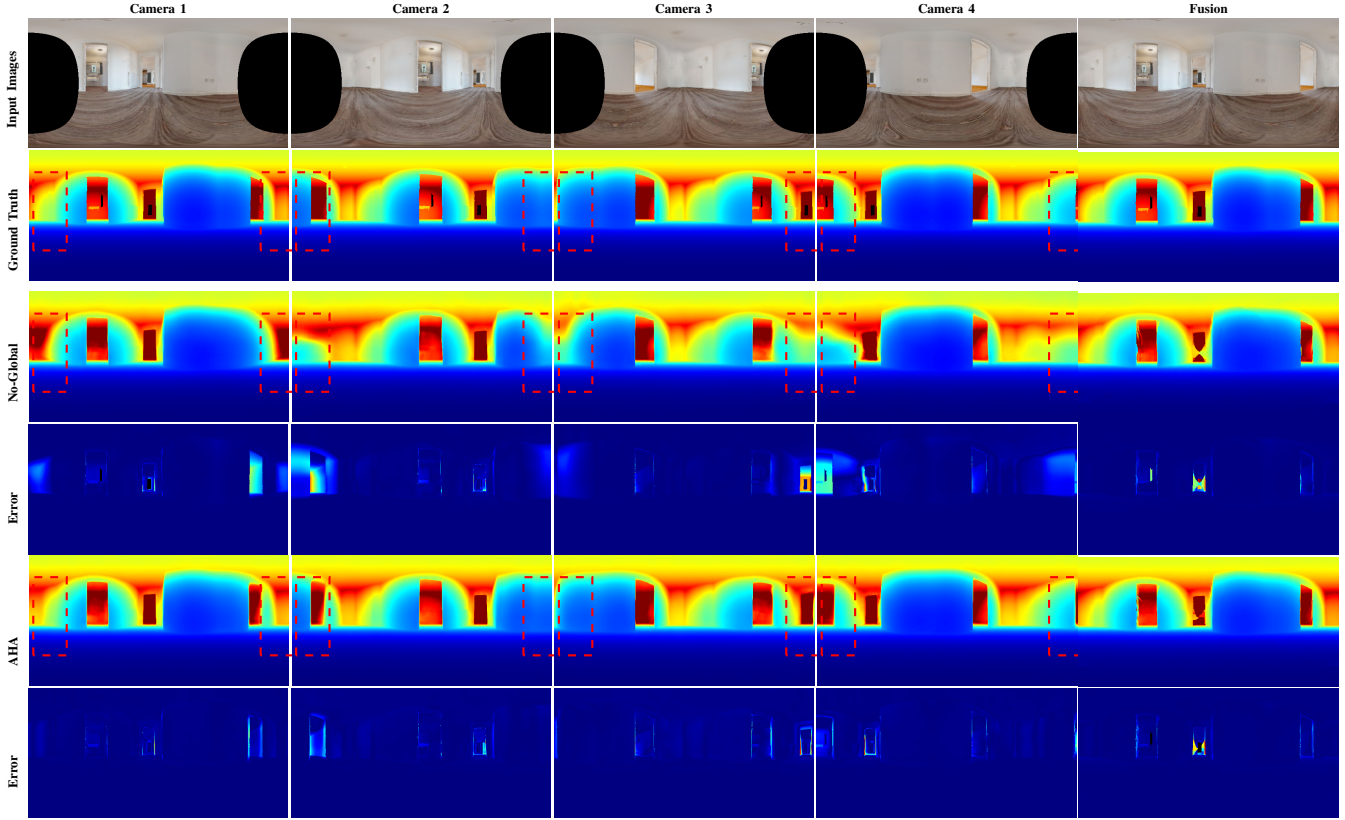

\subsection{ERP fusion}
\label{sec:spherical-fusion}

While our AHA mechanism leverages efficient global attention to achieve good cross-frame/view consistency, we can further enhance depth accuracy through multi-view depth fusion. Traditional fusion methods often struggle with scale 
drift and misalignment issues common in monocular and multi-view depth estimation~\cite{Wang2019Unsupervised, Yang2021D3VO}. 
We propose an ERP fusion approach that projects multi-view depth to a shared ERP coordinate system.

\paragraph{ERP$\rightarrow$3D and pose merging (reference)}
\textbf{Notation.} Let the ERP grid be $W\times H$ with pixel indices $(x,y)\in\{0,\ldots,W{-}1\}\times\{0,\ldots,H{-}1\}$.
For frame $s\in\{1,\ldots,S\}$, $D_s(x,y)$ is the per-pixel depth, $\mathbf{d}(x,y)\in\mathbb{S}^2$ the unit ray recovered from Sec.~\ref{sec:erp}, and $(\mathbf{R}_s,\mathbf{t}_s)$ the extrinsics to a common rig/world frame.
We form 3D points
\[
\mathbf{p}_{s}(x,y) \;=\; \mathbf{R}_s\big(D_s(x,y)\,\mathbf{d}(x,y)\big) + \mathbf{t}_s,
\]
yielding per-frame point sets in a shared coordinate system.

\paragraph{Distance-aware splatting on ERP}
We reproject the points $\{\mathbf{p}_s\}$ to the shared ERP lattice $(u,v)$ and splat each sample at distance $d=\|\mathbf{p}_s\|$ to a $(k\times k)$ window centered at $(u,v)$, where $k=k(d) \in \{1,3,5,7\}$ is monotone in $1/d$ (near--large, far--small). We use \texttt{amin} for depth (z-buffer) and sum/count for color, then apply a light hole filling.
This distance-dependent footprint is inspired by the footprint control in 3D Gaussian Splatting~\cite{kerbl20233dgaussiansplattingrealtime}, adapted to the spherical (ERP) domain.
Let $M_s(u,v)\in\{0,1\}$ denote the per-frame FOV mask on ERP.

\paragraph{Multi-frame ERP fusion (mean)}
On the same ERP grid,
\begin{align}
D_{\text{fuse}}(u,v) &= \frac{\sum_{s=1}^{S} M_s(u,v)\,D_s(u,v)}{\max\!\big(1,\sum_{s} M_s(u,v)\big)},\label{eq:depth-fusion}\\
C_{\text{fuse}}(u,v) &= \frac{\sum_{s=1}^{S} M_s(u,v)\,C_s(u,v)}{\max\!\big(1,\sum_{s} M_s(u,v)\big)}.\label{eq:confidence-fusion}
\end{align}
Empirically, predictions extend slightly beyond nominal FOVs, so masked means provide complementary coverage and reduce inter-view variance.
(Nearest/confidence-weighted variants are supported but omitted for brevity.)

\begin{figure*}[t]
  \centering
    \vspace{0.5em}
  \begin{tabular}{@{}c@{\hspace{-0.6em}}c@{\hspace{-0.6em}}c@{\hspace{-0.6em}}c@{\hspace{-0.6em}}c@{\hspace{-0.6em}}c@{}}
    \begin{tikzpicture}[baseline=(current bounding box.center)]
      \node[anchor=center] (img1) {\includegraphics[width=0.162\textwidth,height=0.12\textwidth,keepaspectratio]{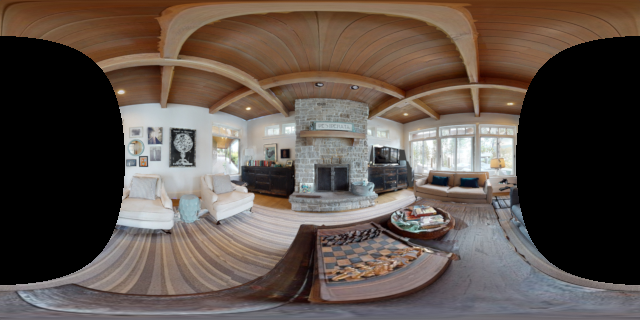}};
    \end{tikzpicture} &
    \begin{tikzpicture}[baseline=(current bounding box.center)]
      \node[anchor=center] (img2) {\includegraphics[width=0.162\textwidth,height=0.12\textwidth,keepaspectratio]{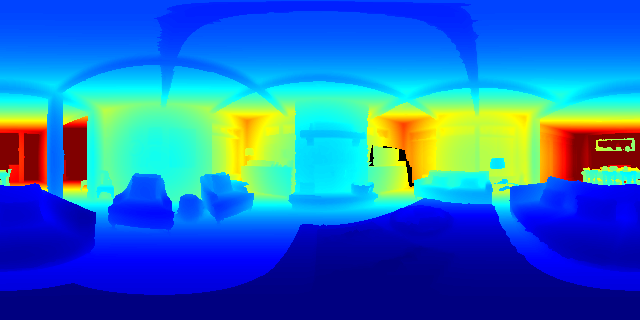}};
      \draw[red, dashed, thick] (img2.south west) ++(0.05*0.169*\textwidth, 0.35*0.169*0.6*\textwidth) rectangle ++(0.125*0.169*\textwidth, 0.35*0.169*0.6*\textwidth);
    \end{tikzpicture} &
    \begin{tikzpicture}[baseline=(current bounding box.center)]
      \node[anchor=center] (img3) {\includegraphics[width=0.162\textwidth,height=0.12\textwidth,keepaspectratio]{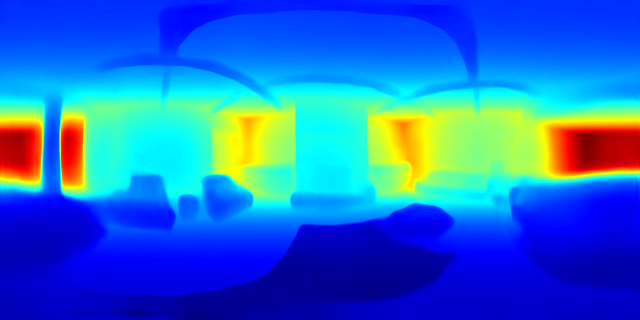}};
      \draw[red, dashed, thick] (img3.south west) ++(0.05*0.169*\textwidth, 0.35*0.169*0.6*\textwidth) rectangle ++(0.125*0.169*\textwidth, 0.35*0.169*0.6*\textwidth);
    \end{tikzpicture} &
    \begin{tikzpicture}[baseline=(current bounding box.center)]
      \node[anchor=center] (img4) {\includegraphics[width=0.162\textwidth,height=0.12\textwidth,keepaspectratio]{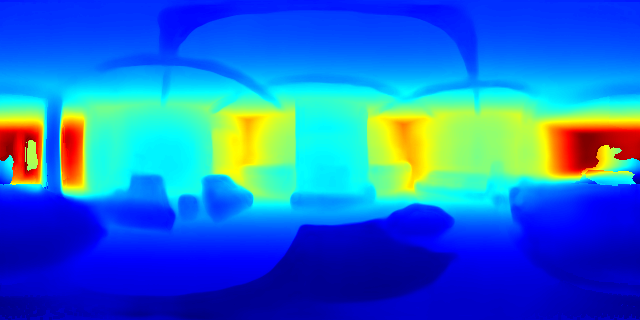}};
    \end{tikzpicture} &
    \begin{tikzpicture}[baseline=(current bounding box.center)]
      \node[anchor=center] (img5) {\includegraphics[width=0.162\textwidth,height=0.12\textwidth,keepaspectratio]{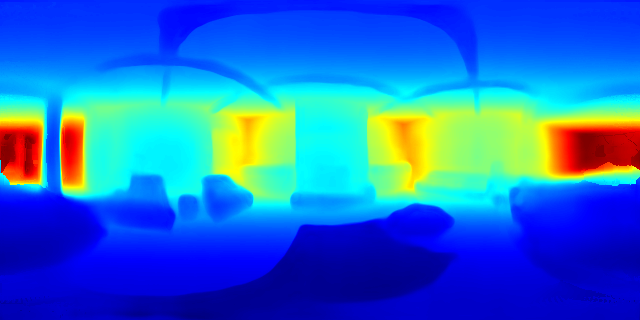}};
    \end{tikzpicture} &
    \begin{tikzpicture}[baseline=(current bounding box.center)]
      \node[anchor=center] (img6) {\includegraphics[width=0.162\textwidth,height=0.12\textwidth,keepaspectratio]{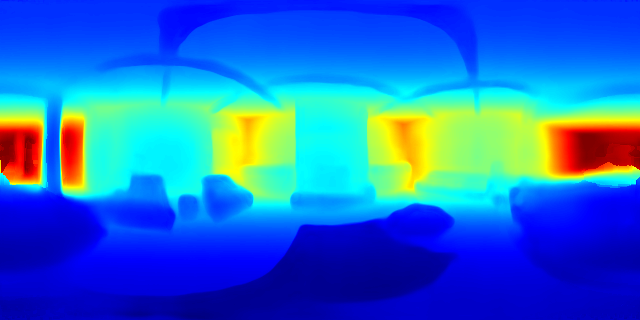}};
      \draw[red, dashed, thick] (img6.south west) ++(0.05*0.169*\textwidth, 0.35*0.169*0.6*\textwidth) rectangle ++(0.125*0.169*\textwidth, 0.35*0.169*0.6*\textwidth);
    \end{tikzpicture}
    \\[-4pt]
    \tiny{(a) Camera 1} & \tiny{(b) GT} & \tiny{(c) No Fusion} & \tiny{(d) Nearest} & \tiny{(e) Weighted} & \tiny{(f) Mean} \\[-4pt]
  \end{tabular}
  
  \caption{Fusion strategy comparison showing different fusion methods. Main differences are highlighted with red dashed rectangles.}
  \label{fig:fusion-comparison}
  \end{figure*}
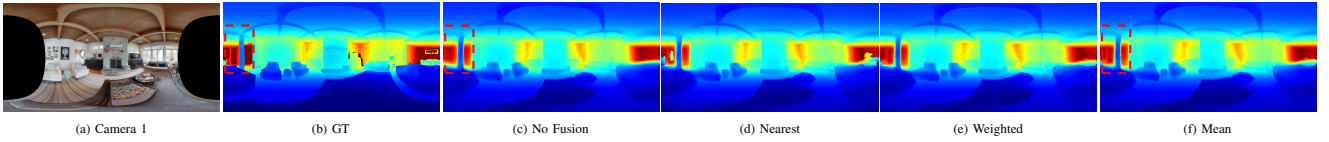

  \begin{figure*}[t]
    \centering
    \setlength{\tabcolsep}{0.1pt}    
    \renewcommand{\arraystretch}{0.294} 
    \begin{tabular}{@{}l@{\hspace{0.1em}}c@{\hspace{0.1em}}c@{\hspace{0.1em}}c@{\hspace{0.1em}}c@{\hspace{0.1em}}c@{\hspace{0.1em}}c@{}}
      \rotatebox{90}{\tiny\textbf{\ \ \ \ \ \ \ \ GT}} &
      \includegraphics[width=0.16\textwidth]{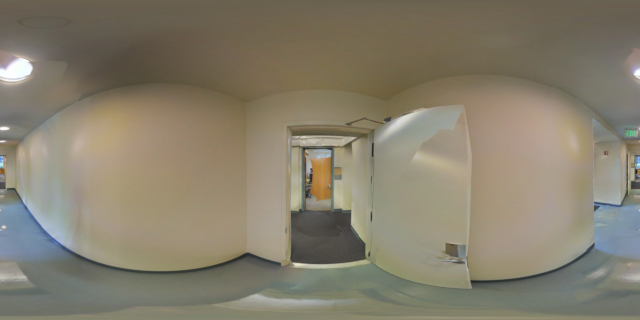} &
      \includegraphics[width=0.16\textwidth]{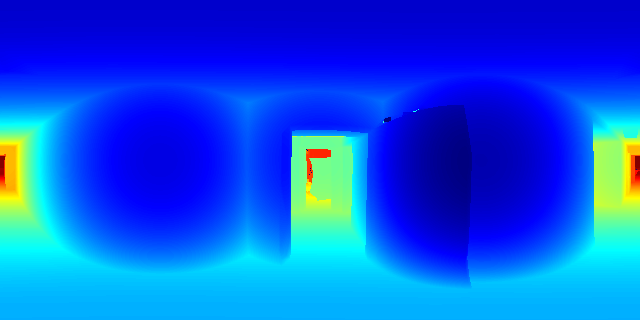} &
      \includegraphics[width=0.16\textwidth]{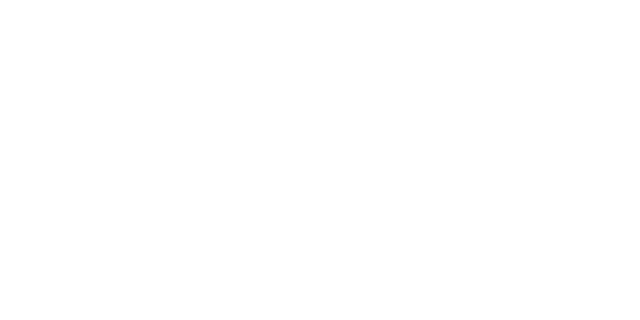} &
      \includegraphics[width=0.16\textwidth]{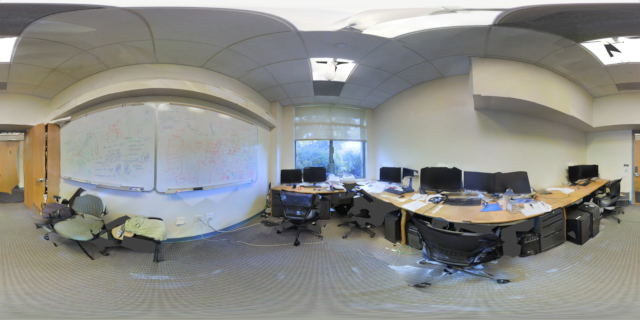} &
      \includegraphics[width=0.16\textwidth]{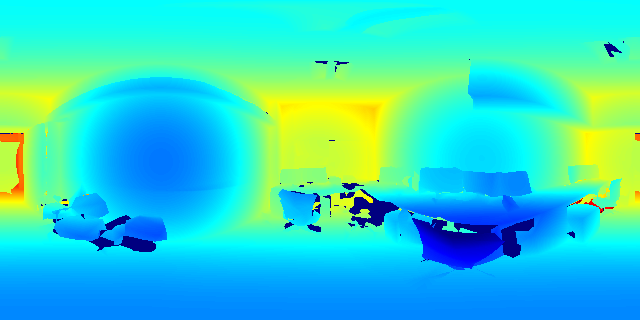} &
      \includegraphics[width=0.16\textwidth]{imgs/zero_shot_on_2d3dsfix/blank.png} \\
      
      \rotatebox{90}{\tiny\textbf{\ \ \ \ \ \ VGGT}} &
      \includegraphics[width=0.16\textwidth]{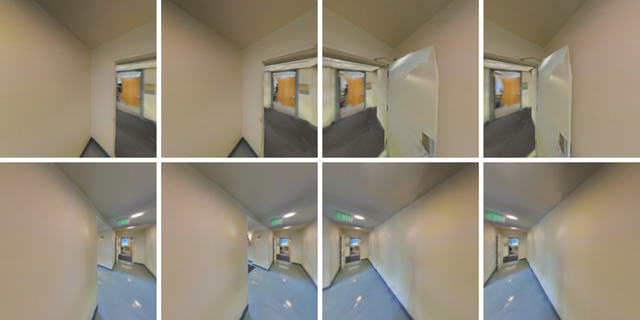} &
      \includegraphics[width=0.16\textwidth]{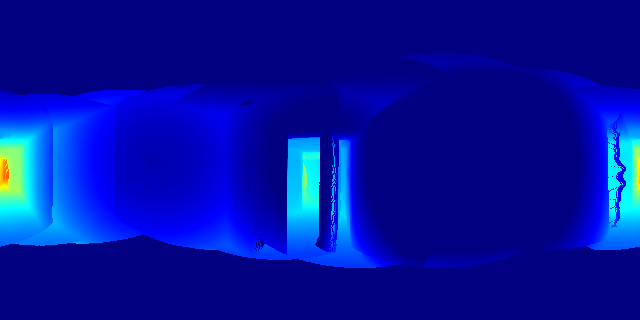} &
      \includegraphics[width=0.16\textwidth]{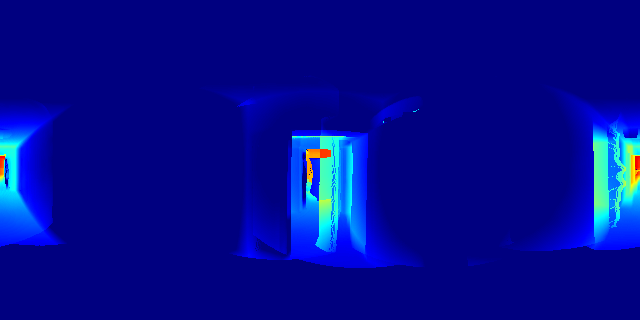} &
      \includegraphics[width=0.16\textwidth]{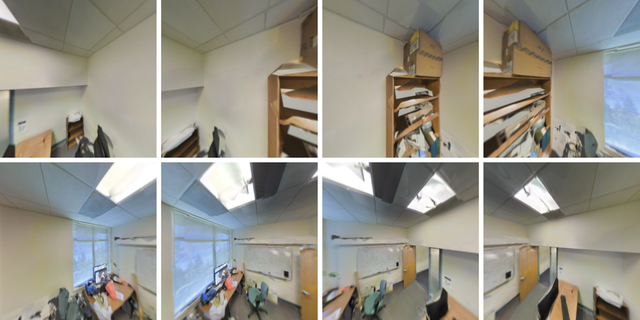} &
      \includegraphics[width=0.16\textwidth]{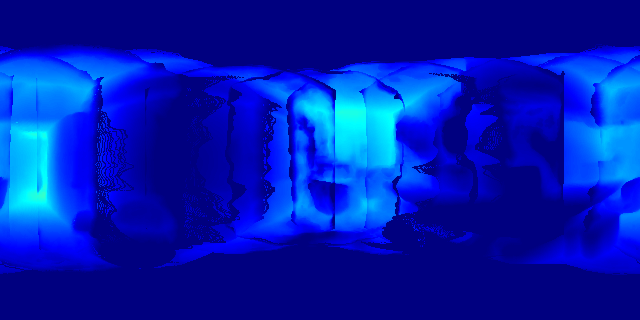} &
      \includegraphics[width=0.16\textwidth]{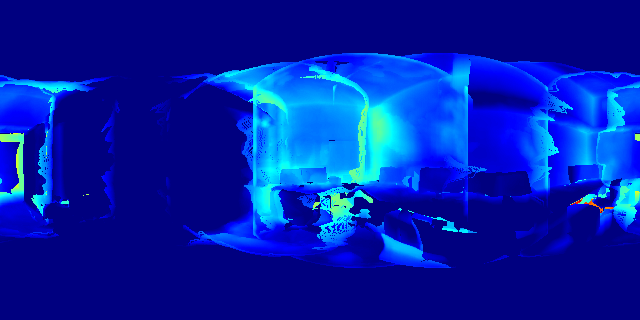} \\
      
      \rotatebox{90}{\tiny\textbf{\ \ \ \ \ OmniStereo}} &
      \includegraphics[width=0.16\textwidth]{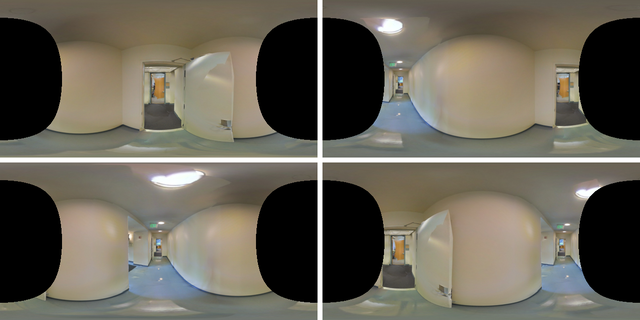} &
      \includegraphics[width=0.16\textwidth]{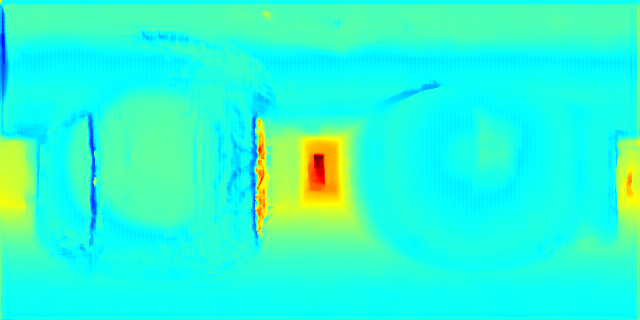} &
      \includegraphics[width=0.16\textwidth]{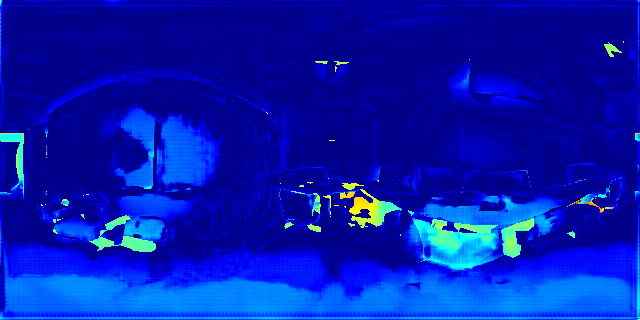} &
      \includegraphics[width=0.16\textwidth]{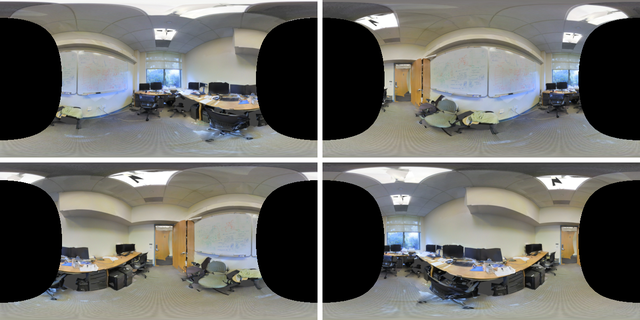} &
      \includegraphics[width=0.16\textwidth]{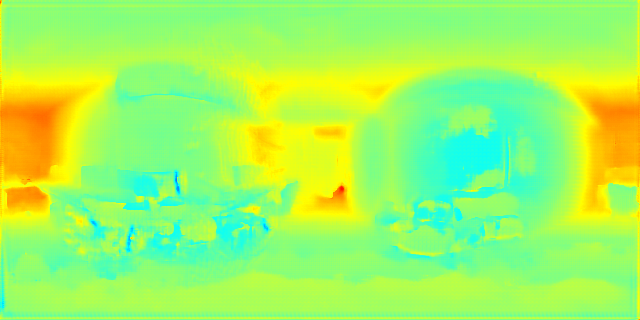} &
      \includegraphics[width=0.16\textwidth]{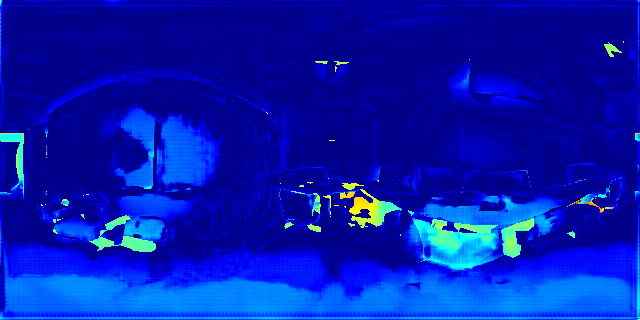} \\
      
      \rotatebox{90}{\tiny\textbf{\ \ \ \ LightStereo}} &
      \includegraphics[width=0.16\textwidth]{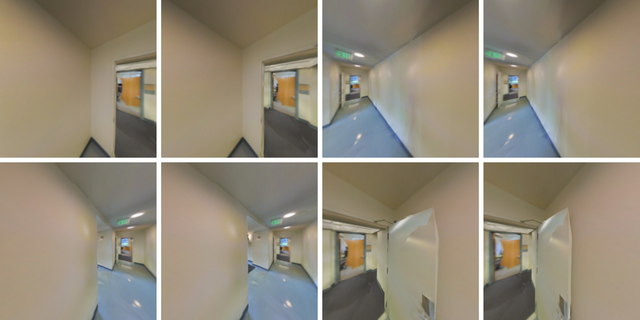} &
      \includegraphics[width=0.16\textwidth]{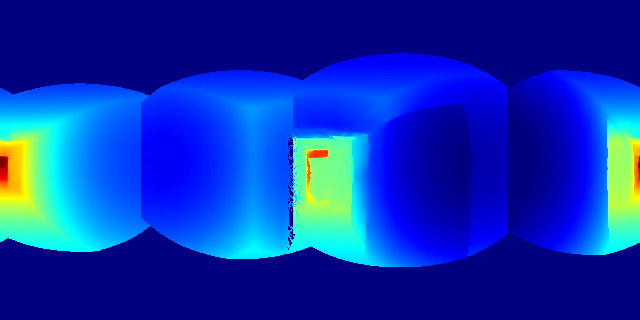} &
      \includegraphics[width=0.16\textwidth]{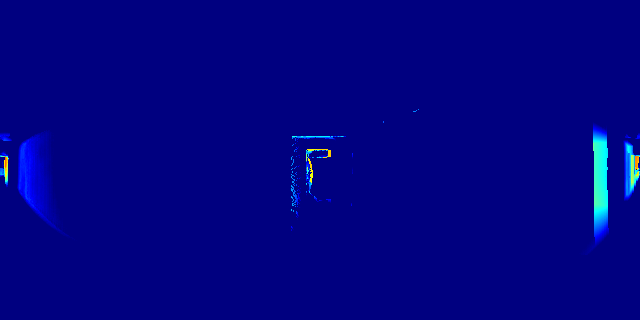} &
      \includegraphics[width=0.16\textwidth]{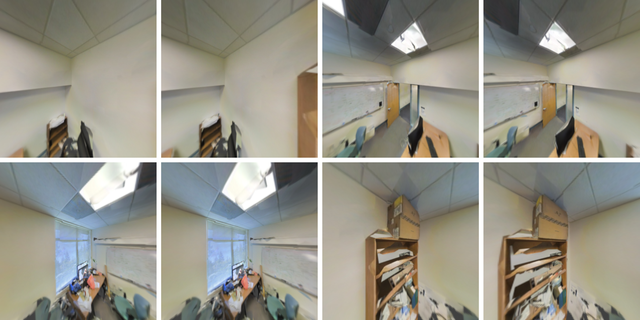} &
      \includegraphics[width=0.16\textwidth]{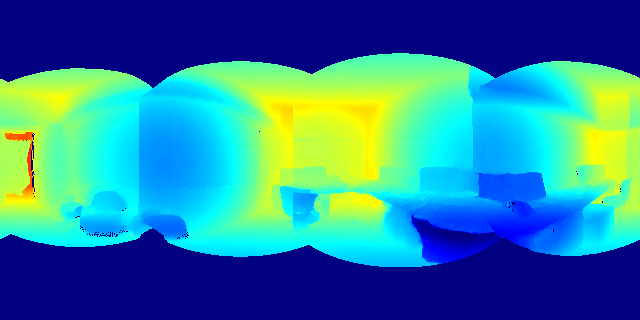} &
      \includegraphics[width=0.16\textwidth]{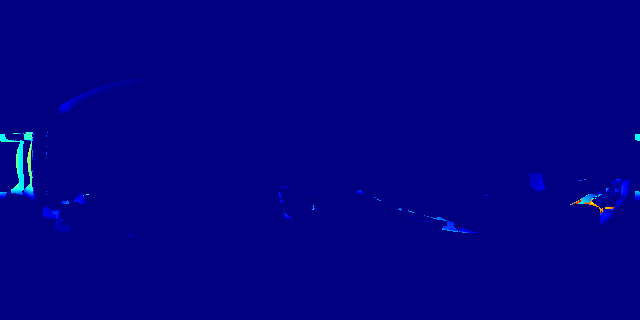} \\
      
      \rotatebox{90}{\tiny\textbf{\ \ \ \ FastViDAR}} &
      \includegraphics[width=0.16\textwidth]{imgs/zero_shot_on_2d3dsfix/erp_1.png} &
      \includegraphics[width=0.16\textwidth]{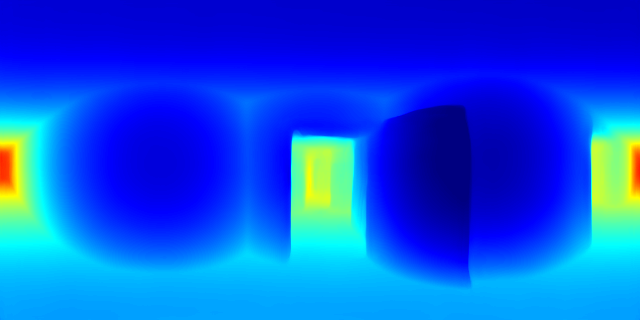} &
      \includegraphics[width=0.16\textwidth]{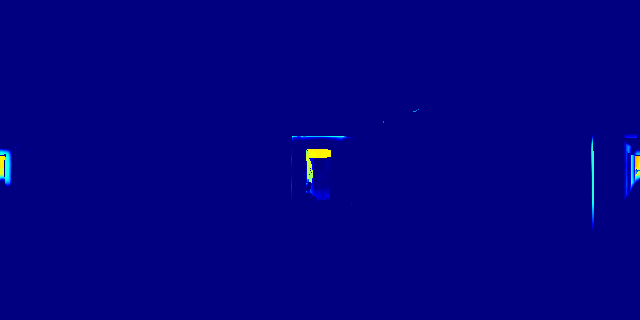} &
      \includegraphics[width=0.16\textwidth]{imgs/zero_shot_on_2d3dsfix/erp_2.png} &
      \includegraphics[width=0.16\textwidth]{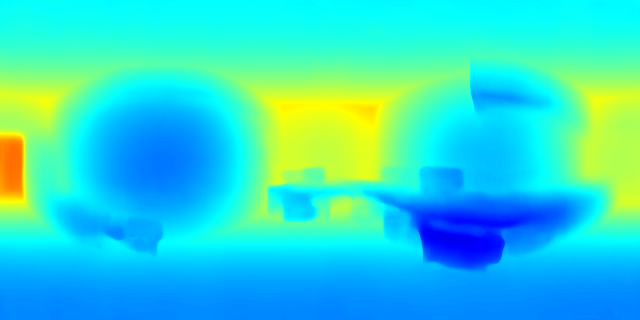} &
      \includegraphics[width=0.16\textwidth]{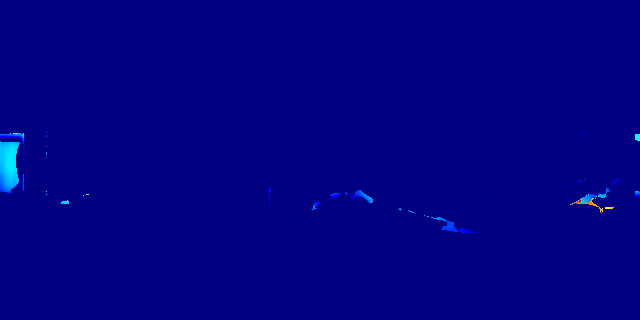} \\[3pt]
      
      & \tiny{{Input}} & \tiny{Pred} & \tiny{Error} & \tiny{Input} & \tiny{Pred} & \tiny{Error}
    \end{tabular}
    \\[-3pt]
    \caption{Comprehensive comparison of FastViDAR with baselines on 2D-3D-S datasets. First row shows the ground truth depth map. Subsequent rows present predictions and error maps for each method. In error maps, regions with $\delta < 1.25$ are shown in pure dark blue.}
    \label{fig:main-visual}
    \end{figure*}

\begin{table}[t]
  \vspace{0.5em}
  \centering
  \caption{Ablation A - Hierarchy. Only global attention on summary tokens is toggled. (w: window, f: frame-level, g: global-level) Best results are highlighted in \textbf{bold}.}
  \label{tab:abl-hierarchy}
  \setlength\tabcolsep{4pt}
  \begin{tabular}{lccccc}
  \toprule
  Variant & AbsRel$\downarrow$ & RMSE$\downarrow$ & Log10$\downarrow$ & $\delta < 1.25$ $\uparrow$ & Time (ms) \\
  \midrule
  No-Global (w+f) & 0.135 & 0.454 & 0.181 & 0.892 & 34 \\
  \textbf{AHA (w+f+g)} & \textbf{0.111} & \textbf{0.384} & \textbf{0.163} & \textbf{0.904} & \textbf{36} \\
  \bottomrule
  \end{tabular}
  \end{table}

  \begin{table}[t]
    \centering
    \caption{Fusion strategies comparison. Best results are highlighted in \textbf{bold}.}
    \label{tab:abl-fusion}
    \setlength\tabcolsep{6pt}
    \begin{tabular}{lcccc}
    \toprule
    Method & AbsRel$\downarrow$ & RMSE$\downarrow$ & Log10$\downarrow$ & $\delta < 1.25$ $\uparrow$ \\
    \midrule
    No-fusion       & 0.109 & 0.369 & 0.149 & 0.897 \\
    Nearest       & 0.113 & 0.384 & 0.153 & 0.898 \\
    \textbf{Weighted}      & \textbf{0.108} & 0.365 & \textbf{0.146} & \textbf{0.901} \\
    \textbf{Mean} & \textbf{0.108} & \textbf{0.364} & \textbf{0.146} & \textbf{0.901} \\
    \bottomrule
    \end{tabular}
    \end{table}

\subsection{ERP-weighted loss function}
\label{sec:loss}

We supervise depth with a \emph{masked, ERP-area-weighted} data term and a \emph{multi-scale gradient} term; both operate per frame and are averaged across $S$ frames. Let $M_s\!\in\!\{0,1\}^{H\times W}$ be the validity mask, $\hat D_s$ the prediction, $D_s$ the target, and (optionally) $C_s\!\ge\!0$ a per-pixel confidence (from the network or set to $\mathbf{1}$). To compensate ERP latitude distortion, rows are weighted by
\[
w(v)=\cos\phi(v),\qquad \phi(v)=\pi\!\big((v+0.5)/H-1/2\big).
\]

\paragraph{Data term}
We use a robust residual with (optional) confidence:
\begin{align}
\mathcal{L}^{(s)}_{\text{data}}
&= \sum_{u,v} w(v)\; M_s(u,v)\, C_s(u,v)\nonumber\\
&\quad \times \rho\!\big(\hat D_s(u,v)-D_s(u,v)\big),\label{eq:loss-data}
\end{align}
where $\rho(\cdot)$ is the Huber loss (we adopt $\delta\!=\!1$ unless noted).

\paragraph{Multi-scale gradient term}
At scales $r=0,\ldots,R{-}1$ (subsample by $2^r$ in $u$ and $v$), we match ERP gradients with the same area weights:
\begin{align}
\mathcal{L}^{(s)}_{\nabla}
&= \frac{1}{R}\sum_{r=0}^{R-1}\;\sum_{u,v} w_r(v)\; M^{(r)}_s(u,v)\, C^{(r)}_s(u,v)\nonumber\\
&\quad \times \rho\!\big(\nabla \hat D^{(r)}_s(u,v) - \nabla D^{(r)}_s(u,v)\big),\label{eq:loss-grad}
\end{align}
with $w_r(v)=\cos\phi_r(v)$ defined analogously at height $H/2^r$, and $\nabla$ the finite-difference operator.

\paragraph{Depth objective and optional regularizer}
Our per-frame depth loss is
\begin{equation}
\mathcal{L}_{\text{depth}} = \frac{1}{S}\sum_{s=1}^{S}
\Big(\mathcal{L}^{(s)}_{\text{data}} + \lambda_{\nabla}\,\mathcal{L}^{(s)}_{\nabla}\Big).
\label{eq:loss-depth}
\end{equation}
The ERP weighting $w(v)$ mirrors our implementation (row-wise $\cos\phi$) and ensures equal-solid-angle supervision; it consistently improves stability by preventing the poles—small area but dense pixels—from dominating the loss.

\begin{table}[t]
\centering
\caption{Performance and efficiency comparison on 2D-3D-S (zero-shot) dataset. Best results are shown in \textbf{bold}.}
\label{tab:main-results}
\setlength\tabcolsep{4pt}
\begin{tabular}{lccccc}
\toprule
Method & AbsRel$\downarrow$ & RMSE$\downarrow$ & Log10$\downarrow$ & $\delta < 1.25$ $\uparrow$ & Time (ms) \\
\midrule
VGGT & 0.557 & 1.934 & 0.396 & 0.043 & 120 \\
OmniStereo & 0.619 & 1.450 & 0.154 & 0.554 & 66 \\
LightStereo & 0.125 & 0.667 & 0.050 & 0.851 & \textbf{33} \\
\textbf{FastViDAR} & \textbf{0.119} & \textbf{0.433} & \textbf{0.046} & \textbf{0.929} & 36 \\
\bottomrule
\end{tabular}
\end{table}

\section{Experiments}
\label{sec:exp}

We evaluate FastViDAR with the AHA backbone (Sec.~\ref{sec:aha}), ERP fusion (Sec.~\ref{sec:spherical-fusion}), and the ERP-weighted loss (Sec.~\ref{sec:loss}). Unless noted, inputs are ERP $640{\times}320$, frame count $S{=}4$, window $(P_h,P_w){=}(7,7)$, and Stage-4 uses $N_4{=}2$ window-MHSA layers. Our experiments use four cameras where each captures one frame, and the network processes all 4 frames simultaneously. We train the model from scratch using AdamW optimizer~\cite{loshchilov2019decoupledweightdecayregularization} with learning rate $1{\times}10^{-4}$, batch size 20, for 20 epochs. We employ OneCycleLR scheduler~\cite{smith2018superconvergencefasttrainingneural} with 10\% warm-up and cosine annealing strategy. For inference time comparison, we test on RTX 4090 with the following input resolutions: LightStereo and VGGT at $4{\times}512{\times}512$, OmniStereo and FastViDAR at $4{\times}640{\times}320$.

\subsection{Datasets and Evaluation Protocol}
\label{sec:exp-data}

\textbf{HM3D (train/ablate).} We render multi-view ERP from 800/200 train/test scenes of HM3D~\cite{HM3D2021}. Each sample uses a 4-camera rigid rig with \emph{randomized} relative poses and FOV in $[160^\circ,360^\circ]$, yielding diverse baselines and overlaps. The dataset contains 421,127 training groups and 52,484 test groups, where each group consists of 4 ERP views.

\textbf{2D-3D-S (zero-shot).} We render a \emph{fixed} rigid ring of 4 fisheye cameras (FOV $220^\circ$) with baseline $20\sqrt{2}$\,mm and $90^\circ$ angular separation between adjacent cameras from 6 large scenes, totaling 6,000 groups. Each camera captures one frame, and the network processes all 4 frames simultaneously. No fine-tuning is performed on 2D-3D-S~\cite{Armeni2017S2D3DS}.

\textbf{Training data.} For ablation experiments, our method is trained only on images collected from the 800 training scenes of HM3D. For zero-shot evaluation on 2D-3D-S, our method is trained on the complete HM3D dataset with 1,000 scenes plus 200 publicly available Blend scenes (excluding 3D models from 2D-3D-S), with no additional data used.

\textbf{Preprocessing \& evaluation domain.} All fisheye views are mapped to a common ERP lattice (Sec.~\ref{sec:erp}). All metrics are computed on the ERP grid using the \emph{intersection} validity mask $M(u,v)=M_{\text{gt}}(u,v)\wedge M_{\text{meth}}(u,v)$ (i.e., GT-valid $\cap$ method-available FOV). For fairness, all depth predictions are considered regardless of confidence values. All multi-view depth predictions are transformed to the first view's coordinate system before evaluation.

\subsection{Metrics}
\label{sec:exp-metrics}

All scores are computed on the ERP grid under the intersection mask $M$ defined above. Let $Z=\sum_{u,v} M(u,v)$ and define the masked mean
\begin{equation}
\mathbb{E}_M[f] \triangleq Z^{-1}\!\sum_{u,v} M(u,v)\,f(u,v).
\label{eq:masked-mean}
\end{equation}
Given ground-truth depth $D$ and prediction $\hat D$ (in meters), set $\Delta=\hat D-D$ and fix $\varepsilon>0$.

\paragraph{Unified error functionals (lower is better)}
For a transform $g:\mathbb{R}_{+}\!\to\!\mathbb{R}$ and $p\!\ge\!1$,
\begin{align}
\mathcal{L}_{p}(g) &\triangleq \Big(\mathbb{E}_M\!\big[\,|g(\hat D)-g(D)|^p\,\big]\Big)^{1/p},
\label{eq:error-functional-L}\\
\mathcal{R}_{q} &\triangleq \mathbb{E}_M\!\Big[\,\big(\tfrac{|\Delta|}{D+\varepsilon}\big)^{q}\,\Big], \quad q\!\ge\!1.
\label{eq:error-functional-R}
\end{align}
For threshold accuracy (higher is better), let $\tau=\max(\hat D/D,\;D/\hat D)$ and define
\begin{equation}
\mathrm{A}(\alpha)=\mathbb{E}_M\big[\mathbf{1}(\tau<\alpha)\big].
\label{eq:threshold-accuracy}
\end{equation}

\noindent\textbf{Reported metrics.}
We report: \textbf{AbsRel}$=\mathcal{R}_1$, \textbf{RMSE}$=\mathcal{L}_2$ with $g(x)=x$, \textbf{Log10}$=\mathcal{L}_1(\log_{10})$, and $\boldsymbol{\delta<1.25}=\mathrm{A}(1.25)$.

\paragraph{Protocol}
Scores are computed per scene (the same $M$ for all methods within a scene) and averaged over the test split. Unless stated, $\varepsilon=10^{-6}$. RMSE is reported in meters; other metrics are dimensionless.

\subsection{Ablation Studies and Analysis}
\label{sec:exp-abl}

We conduct ablation studies focusing on our two key contributions—\emph{AHA} and \emph{ERP fusion}—under identical training and evaluation settings.

\paragraph{AHA vs No-Global} We disable global attention on summary tokens (keeping window and per-frame attention). Capacity, loss, and training schedule remain unchanged. Table~\ref{tab:abl-hierarchy} shows that AHA improves accuracy metrics while maintaining efficiency, confirming that summary-level global reasoning improves cross-frame consistency. Qualitative results are shown in Figure~\ref{fig:abl-visual}. AHA demonstrates improved performance with better scale accuracy and consistency, particularly in regions outside each camera's FOV, providing more accurate depth predictions and richer details compared to the No-Global attention baseline.

\paragraph{Fusion strategy} We compare no-fusion (per-frame), mean (ours), nearest, and confidence-weighted fusion strategies. Table~\ref{tab:abl-fusion} demonstrates that mean fusion consistently achieves the best performance, providing stable and accurate depth predictions across different scenarios. Qualitative comparison results are shown in Figure~\ref{fig:fusion-comparison}.

\subsection{Zero-Shot Comparison with State-of-the-Art Methods}
\label{sec:exp-main}

We evaluate FastViDAR against state-of-the-art baselines: (i) VGGT (MVS), (ii) OmniStereo (omnidirectional depth), and (iii) LightStereo-M (real-time stereo). VGGT and OmniStereo use pre-trained weights without adaptation, while LightStereo-M is fine-tuned on our training data. LightStereo-M has been extensively trained on over 1.5 million samples from FoundationStereo~\cite{wen2025foundationstereozeroshotstereomatching} and other datasets~\cite{mayer2016sceneflow,li2022crestereo,tremblay2018fallingthings,butler2012sintel,cabon2020vkitti2,dosovitskiy2017carla,geiger2012kitti12,menze2015kitti15,hirschmuller2007middlebury0506,scharstein2014middlebury,pan2021middlebury,scharstein2002middleburyQH,schops2017eth3d,bao2020instereo2k,ramirez2022booster}. For LightStereo, we split ERP into stereo pairs and convert disparity to depth. For VGGT, we split 4 ERP images into 8 directional pinhole views ($100^\circ$ FOV) and project output point clouds to ERP depth maps. OmniStereo and FastViDAR use the original 4 ERP images ($220^\circ$ FOV).

Table~\ref{tab:main-results} presents zero-shot results on the 2D-3D-S dataset. FastViDAR achieves competitive performance across all metrics without fine-tuning, demonstrating strong cross-domain generalization. Qualitative results are shown in Figure~\ref{fig:main-visual}. Although slightly slower than LightStereo, our method provides comprehensive $360^\circ$ depth coverage with better multi-view consistency and accuracy.

\section{Conclusion} 
\label{sec:exp-conclusion}
We propose FastViDAR, a real-time omnidirectional multi-view depth estimation method featuring Alternative Hierarchical Attention (AHA) and ERP fusion. Despite being trained on only 1,200 scenes, our method shows competitive results compared to OmniStereo, VGGT, and LightStereo (trained on 1.5M+ samples), demonstrating the effectiveness of AHA and fusion methods. Our approach provides improved depth consistency and accuracy across all spatial directions, with flexible pose handling that supports arbitrary camera configurations, making it suitable for applications requiring comprehensive $360^\circ$ depth perception.

\newpage
\bibliographystyle{IEEEtran}
\bibliography{refs}

\end{document}